\documentclass{article}

\usepackage{amsmath, amsfonts, amssymb, amsthm}
\usepackage{txfonts} 
\usepackage{verbatim}
\usepackage{url}
\usepackage{color}
\usepackage{algorithm2e}
\usepackage[noend]{algpseudocode}
\usepackage{multirow}

\usepackage{siunitx}
\ifx\BlackBox\undefined
\newcommand{\BlackBox}{\rule{1.5ex}{1.5ex}}  
\fi
\ifx\proof\undefined
\newenvironment{proof}{\par\noindent{\bf Proof\ }}{\hfill\BlackBox\\[2mm]}
\fi

\newcommand\shortsection[1]{\vspace{6pt}{\noindent\bf #1.}}
\newcommand\shortersection[1]{\vspace{6pt}{\noindent\em #1.}}

\newtheorem{theorem}{Theorem}[section]

\newtheorem{lemma}[theorem]{Lemma}
\theoremstyle{definition}

\makeatletter
\def\url@leostyle{%
  \@ifundefined{selectfont}{\def\UrlFont{\sf}}{\def\UrlFont{\small\sffamily}}}
\makeatother
\makeatletter
\def\url@beostyle{%
  \@ifundefined{selectfont}{\def\UrlFont{\sf}}{\def\UrlFont{\scriptsize\sffamily}}}
\makeatother

\newcommand{\redacted}[1]{{\emph{$<$Redacted for Anonymity$>$}}}

\usepackage[margin=1in]{geometry}

\usepackage{tikz}
\usepackage{amsmath}

\usepackage{amsthm}
\usepackage{subcaption}
\usepackage{nicefrac}       
\usepackage{microtype}
\usepackage{graphicx}
\usepackage{booktabs}       
\usepackage{caption}
\theoremstyle{definition}
\usepackage{dashbox}
\usepackage[
    n,
    operators,
    advantage,
    sets,
    adversary,
    landau,
    probability,
    notions,    
    logic,
    ff,
    mm,
    primitives,
    events,
    complexity,
    asymptotics,
    keys]{cryptocode}
\usepackage{amssymb}
\usepackage{caption}
\usepackage{cleveref}
\usepackage{multirow}

\newcommand{\boneage}{RSNA Bone Age}
\newcommand{\graph}{ogbn-arxiv}
\newcommand{\census}{Census}
\newcommand{\celeba}{CelebA}
\newcommand{\botnet}{Chord}
\newcommand{\ltest}{Loss Test}
\newcommand{\ttest}{Threshold Test}

\newcommand{\acc}{\textit{acc}}
\newcommand{\neff}{\ensuremath{n_\textrm{leaked}}}

\newcommand{\gzero}{$\mathcal{G}_0$}
\newcommand{\gone}{$\mathcal{G}_1$}

\begin{document}

\urlstyle{leo}
\date{}

\title{\Large \bf Formalizing and Estimating Distribution Inference Risks}

\author{
Anshuman Suri and David Evans \\
Department of Computer Science\\
University of Virginia\\
\textsf{\small \{anshuman, evans\}@virginia.edu}
} 

\maketitle

\begin{abstract}
Distribution inference, sometimes called property inference, infers statistical properties about a training set from access to a model trained on that data. Distribution inference attacks can pose serious risks when models are trained on private data, but are difficult to distinguish from the intrinsic purpose of statistical machine learning---namely, to produce models that capture statistical properties about a distribution. 
Motivated by Yeom et al.'s membership inference framework, we propose a formal definition of distribution inference attacks that is general enough to describe a broad class of attacks distinguishing between possible training distributions.
We show how our definition captures previous ratio-based property inference attacks as well as new kinds of attack including revealing the average node degree or clustering coefficient of a training graph. 
To understand distribution inference risks, we introduce a metric that quantifies observed leakage by relating it to the leakage that would occur if samples from the training distribution were provided directly to the adversary. We report on a series of experiments across a range of different distributions using both novel black-box attacks and improved versions of the state-of-the-art white-box attacks. Our results show that inexpensive attacks are often as effective as expensive meta-classifier attacks, and that there are surprising asymmetries in the effectiveness of attacks.

\end{abstract}

\section{Introduction}
\label{sec:introduction}

Inference attacks aim to infer sensitive information about inputs to a process from its revealed outputs. In machine learning, such attacks usually focus on learning sensitive information about training data from a released model. For example, in a \emph{membership inference} attack~\cite{shokri2017membership}, the adversary aims to infer whether a particular record was part of the training data. 
In a \emph{distribution inference} attack, an adversary aims to infer some statistical property of the training dataset, such as the proportion of women in a dataset used to train a smile-detection model~\cite{ateniese2015hacking}. In the research literature, such attacks have previously been called \emph{property inference} and \emph{attribute inference} (confusingly, since this is also used to refer to a type of dataset inference where the adversary infers an unknown sensitive feature of records in the training dataset~\cite{fredrikson2014privacy}), and various other terms.


The privacy threat posed by membership inference attacks is well recognized---if an adversary can infer the presence of a particular user record in a training dataset of diabetes patients, it would violate privacy laws limiting medical disclosure. Distribution inference attacks pose a less obvious threat but can also be dangerous. As one example, consider a financial organization that trains a loan scoring model on some of its historical data. An adversary may use a distribution inference attack to infer the proportion of the training data having a specific value for some protected attribute (e.g., race), which might be a sensitive property of the training dataset. Distribution inference attacks can also pose a threat to distributed training: curious users wanting to learn sensitive information about training distributions of fellow participants, which are competitors in settings like cross-silo federated learning~\cite{kairouz2021advances}, thus leaking sensitive information. Other examples include inferring the sentiment of emails in a company from a spam classifier, or inferring the volume of transactions from fraud detection systems~\cite{saeed}. We also demonstrate examples where adversaries learn properties of graphs used in training, such as accurately estimating their average clustering coefficient.
Inferring statistical properties of a training distribution can also be used to enhance membership inference attacks or to reveal that a model was trained on a biased dataset~\cite{zhou2021property}.

Previous works have used several different informal notions of property inference attacks (e.g., \cite{gopinath2019property,jegorova2021survey,zhang2021leakage}), but there is no established general formal definition of distribution inference. In this work, we formalize distribution inference attacks based on a critical insight: the key difference between these attacks and other inference attacks is that the adversary's goal in the former is to learn about the training \emph{distribution}, not about the specific training \emph{dataset}. Dataset inference attacks, such as membership inference~\cite{shokri2017membership}, attribute inference~\cite{fredrikson2014privacy}, and ownership-resolution~\cite{maini2021dataset} operate on the level of training records. Attacks like membership inference are directly connected to differential privacy which bounds the ability to distinguish neighboring datasets. By contrast, distribution inference attacks attempt to learn statistical properties of the underlying distribution from which the training dataset is sampled. Having a formal definition and a clear threat model can be useful in several ways---quantifying information leakage, assessing and comparing the practicality of different threat models, and drawing possible links between distribution inference risk and other distribution-level properties such as robustness and fairness.



\shortsection{Contributions}
We provide a general and straightforward formalization of distribution inference attacks as a cryptographic game (Section~\ref{sec:property_defns}), inspired by Yeom \textit{et al.}'s membership inference definition~\cite{yeom2018privacy}. Our definition is generic enough to capture a variety of statistical properties of the underlying distribution including, but not limited to, the attribute ratios that have been the focus of previous property inference attacks.
Using our definition, we define an intuitive metric for measuring the amount of leakage observed in a simulated attack, along with theorems that compute this metric for the case of ratios of binary functions, regressions that estimate the proportion of training data having a given attribute, and degree distributions (for graphs) as properties (Section~\ref{sec:quantifying}).

We report on experiments evaluating distribution inference as a risk across several datasets and properties, providing the first systematic evaluation of how inference risks vary as the distributions diverge. Studying patterns in distribution inference risk and how they correlate with underlying properties helps compare trends across experiments and properties, helping identify datasets and models that are highly sensitive to such inference attacks.
To conduct our experiments we introduce two simple black-box distribution inference attacks, and extend the white-box permutation-invariant network~\cite{ganju2018property} architecture, the current state-of-the-art property inference method, to add support for convolutional layers (Section~\ref{sec:attacks}). This enables us to conduct distribution inference attacks on deep neural networks, even when target models are trained from scratch.
Section~\ref{sec:experiments} reports results from our experiments with these attacks on a variety of datasets, tasks, and inference goals, revealing that simple black-box attacks often perform surprisingly well, and in some settings white-box meta-classifier attacks can reveal information comparable to sampling dozens of records from the training distribution.
We analyze the information gleaned up by meta-classifiers across layers, finding that most information can be find in just one or a few layers, enabling less expensive meta-classifier attacks (Appendix~\ref{sec:layer_imp}). 

\section{Previous Work}
\label{sec:background}

Here we summarize prior work on formal definitions of privacy, distribution inference attacks and proposed defenses.

\shortsection{Privacy definitions}
Most formal privacy definitions, including numerous variations on differential privacy~\cite{desfontaines2020sok}, focus on bounding inferences about specific data elements, not the statistical properties of a dataset. The key privacy notion of traditional differential privacy is intuitively connected to the risk to an individual in contributing their data to a dataset --- this corresponds well to dataset privacy risks, but does not capture distribution inference risks; indeed, the main goal of most differentially private mechanisms is to satisfy the inference bound for individual data while providing the most accurate aggregate statistics possible. One notable exception is the Pufferfish framework~\cite{kifer2014pufferfish}, which introduces notions that allow capturing aggregates of records via explicit specifications of potential secrets (e.g., distribution of vehicle routes in a shipping company) and their relations. Zhang et al.\ extend the Pufferfish framework to define the concept of ``attribute privacy''~\cite{zhang2020attribute}, including a notion of distributional attribute privacy that takes a hierarchical approach for parameterizing distributions and could be instantiated to capture notions of distribution inference such as the fraction of records with some attribute. Although these definitions are promising and valuable, none of them are able to satisfy the simple goal we have to define distribution inference attacks in a way that is general and powerful, while clearly distinguishing inferences that are considered attacks from allowable statistical inferences.

A recent attempt to formalize property inference~\cite{saeed} consists of a framework that reduces property inference to Boolean functions of individual members, posing the ratio of members satisfying the given function in a dataset as the property. These ratio-based formulations limit the kinds of distribution inferences considered since they cannot capture many other kinds of statistical properties of the training distribution that may be sensitive, like the degree distribution of a graph~\cite{hay2009accurate}. Ratio-based formulations assume the property function is applicable over individual data points (with the ratio as the property), while for graphs it is an aggregation over interconnected nodes.

\shortsection{Distribution inference attacks}
All previous distribution inference attacks in the literature take a meta-classifier approach---the adversary trains models on datasets with different properties, then trains a meta-classifier using representations of those models. The adversary then uses the meta-classifier to look at the victim's model and predict a property of the model's training data, which is usually related to the ratio of members satisfying some Boolean property. Ateniese et al.~\cite{ateniese2015hacking} were the first to identify the threat of distribution inference (which they called \emph{property inference}) and proposed a meta-classifier attack targeting Support Vector Machines and Hidden Markov Models. The proposed attacks can successfully infer the accent of speakers in speech-to-text systems, or the presence of traffic from particular sources in network traffic classification systems.
Model representations for training the meta-classifier can take several forms: using model weights themselves ~\cite{ganju2018property}, or model activations or logits for a 
generated set of query points~\cite{xu2019detecting}. These methods show promise, achieving better-than-random results for several properties, various tasks, and models across different domains. For instance, predicting a doctor's specialty based on rating-prediction systems on text reviews~\cite{zhang2021leakage}, identifying accents of speakers in voice-recognition models~\cite{ateniese2015hacking}, and even predicting if a model has been trained to have a backdoor Trojan~\cite{xu2019detecting}. Although these approaches achieve high accuracy rates on ``toy-like" classifiers like decision trees and shallow neural networks, and with distributions that are highly disparate, successful property inference attacks have not been demonstrated on realistic (or even semi-realistic) deep neural networks or complex datasets. Our work is the first to demonstrate the capability of such attacks to work on large convolutional networks and different datasets across domains.

Zhou et al.\ \cite{zhou2021property} extend distribution inference attacks to directly infer property ratios for Generative Adversarial Networks (GANs). Their approach involves the adversary training shadow GANs and then launching a black-box attack on the victim model using generated samples. Although their attack setting includes targeting large GAN models on complex datasets like CelebA~\cite{liu2018large}, the adversary does not use the victim model's model parameters directly as is done by our extension to convolutional models (Section~\ref{sec:attack_conv}). Similarly, Pasquini et al.~\cite{pasquini2021unleashing} extend distribution inference attacks to a split-learning setting and only target parts of the victim model. Zhang et al.\ extended this approach to graphs and their properties~\cite{zhang2021inference}, distinguishing training graphs according to properties such as the number of nodes and edges using attacks that assume access to graph embeddings.
Recent attacks on hyper-parameter stealing~\cite{wang2018stealing} along with practical attacks on large language models~\cite{carlini21extracting} provide further evidence that deep neural networks leak many kinds of information about their training data and process.

\shortsection{Defences against distribution inference}
Unlike other notions of privacy like membership privacy, there are no known defenses against distribution inference. Differential privacy does not mitigate distribution inference risks since it obfuscates the contribution of individual records, while the adversary in our setting cares about statistical properties of the underlying distribution. Attempts in previous works to evaluate differential privacy as a potential defense~\cite{ateniese2015hacking} show that it fails to mitigate distribution inference. Removing sensitive attributes from datasets is not an effective defense either~\cite{zhang2021leakage}, since correlations between attributes still leak information. Using node multiplicative transformations~\cite{ganju2018property} has been proposed as a defense for neural networks with ReLU activations, but provides no protection against black-box attacks.

\section{Distribution Inference}
\label{sec:property_defns}

\shortsection{Threat model}
We model the adversary’s knowledge of the underlying distribution through data sampled from that distribution. For complex, high-dimensional non-synthetic data, this is usually the only way to capture knowledge of a data distribution. While the threat model focuses on distributions, we use actual non-overlapping sampled data to empirically model those distributions.

A natural extension of our threat model incorporates a poisoning opportunity where the adversary can participate in the training process itself. Such an adversary may poison the training dataset by injecting adversarially crafted datapoints (explored by Chase et al.~\cite{saeed}) or control the training procedure itself to introduce some Trojan in the model. Recent works look at a similar scenario where the adversary participates in the learning process via a federated-learning setup, launching attribute-reconstruction attacks using epoch-averaged model gradients~\cite{lyu2021novel}. In this paper, we only consider adversaries with no ability to observe or influence the training process. 

\shortsection{Setup} Let $\mathcal{D}=(\mathcal{X}, \mathcal{Y})$ be a public distribution between data, $\mathcal{X}$, and its corresponding labels, $\mathcal{Y}$. We assume both the model trainer $\mathcal{T}$ and adversary $\mathcal{A}$ have access to $\mathcal{D}$. Both parties also have access to two functions, \gzero\ and \gone, that transform distributions. Inferring properties of the distribution can reveal sensitive information in many scenarios, which can be captured by suitable choices of \gzero\ and \gone.
Using such functions along with the underlying distribution $\mathcal{D}$ makes the setup less restrictive than defining two arbitrary distributions---since the functions \gzero, \gone, and $\mathcal{D}$ are considered public knowledge, anyone can recreate these distributions. Using functions on the same distribution $\mathcal{D}$ emphasizes the fact that these two distributions stem from the same underlying distribution $\mathcal{D}$, enabling the definition to capture a wide class of possible attack goals and scenarios by selecting appropriate functions for \gzero\ and \gone.

To illustrate these definitions, we present a concrete example inspired by Chase et al.~\cite{saeed}. Let $\mathcal{D}$ be a distribution of emails with labels for spam/ham. \gzero\ is applied over $\mathcal{D}$ to yield a modified distribution \gzero($\mathcal{D}$) with 0.8 probability of sampling an email that has negative sentiment, \textit{i.e.} a dataset sampled uniformly at random from this distribution would have approximately 80\% of the emails in it with negative sentiment. Similarly, \gone($\mathcal{D}$) could be another distribution with this probability as 0.5 (equally likely to be positive or negative). The adversary thus wants to know if the training distribution is biased, which, if inferred near the financial quarter, can be used to predict if the company is performing below expectations. Alternatively, an ambitious adversary could even consider directly inferring this proportion (which was not considered in Chase et al.~\cite{saeed}, but we explore in Theorem~\ref{thm:regress_neff} and our regression experiments on other datasets).

We propose a general and straightforward experiment to formalize property inference attacks, inspired by Yeom \textit{et al.}'s cryptographic game definition of membership inference~\cite{yeom2018privacy}.
In our cryptographic game definition, $\mathcal{T}$ picks one of the $\mathcal{G}_{\{0, 1\}}$ distribution transformers at random and samples a dataset $S$ from the resulting distribution.
Given access to a model $M$ trained on $S$, the adversary aims to infer which of the two distribution mappers was used:
\begin{figure}[h]
    \centering
    \fbox{%
        \pseudocode[]{%
            \textbf{Trainer } \mathcal{T} \<\< \textbf{Adversary } \adv \\[][\hline]
             \\
            \pcln b \sample \bin \\
            \pcln S \sim \mathcal{G}_b(\mathcal{D}) \\
            \pcln M \xleftarrow{train}{} S  \\
            \pcln \< \sendmessageright{top={$M$}, length=2cm} \\
            \pcln \< \< \hat{b} = \mathcal{H}(M)
        }
    }
\end{figure}

In this work, we assume the adversary has no control over the training process (Step 3). For cases where the adversary has access to the training data, it can trivially infer desired properties by inspecting it. Our definition could be adapted to other scenarios such as federated learning~\cite{li2020federated} by providing additional information to the adversary or allowing the adversary to have some control over $S$ or other aspects of the training process, but we do not consider such settings in this paper.

If $\adv$ can successfully predict $b$ via $\hat{b}$, then it can determine which of the training distributions was used. The advantage of the adversary $\adv$ using algorithm $\mathcal{H}$ is defined as:
$$ \text{Adv}_{\mathcal{H}} = \abs{\condprob{\hat{b}}{b} - \condprob{\hat{b}}{\neg\ b}}.
$$
This advantage is negligible when the adversary does no better than random guessing.
We do not assume the adversary knows how training data is collected, or has any access to it: just that it knows the underlying common distribution $\mathcal{D}$, and has a goal of distinguishing between sub-distributions \gzero($\mathcal{D}$), \gone($\mathcal{D}$) of that distribution. The adversary does not need to know the actual training distribution---indeed, learning about this is the goal of the attack. They just need to have hypotheses worth testing, and thus \gzero\ and \gone\ are defined by the adversary based on what they want to test.

\shortsection{Limitations of the Definition}
This definition is simple and general, but does not capture all kinds of distribution inference attacks. It assumes a setting where the adversary attempts to distinguish between two particular distributions, both of which are defined by the adversary. Multiple experiments could extend the definition to a set of possible distributions, but the definition does not directly capture the regression attacks we demonstrate where the adversary is estimating what proportion of a training dataset has a given property directly.
Our definition also assumes the adversary has prior knowledge of the two sub-distributions to distinguish. Some knowledge of the statistical property the adversary wants to learn about the victim's training distribution is inherent in the nature of a distribution inference attack, but  this may not always be in the form of knowledge of possible distributions. In our experiments, we model an adversaries knowledge of the underlying distribution through a sampled, non-overlapping dataset, which the adversary may then use to construct approximations of different sub-distributions.


\shortsection{Applying the Definition}
\label{sec:existing_defns}
Seminal works on property inference~\cite{ateniese2015hacking,ganju2018property} involve a model trained either on the original dataset or a version modified to be biased towards some chosen attribute. Our definition can be used to describe these attacks by setting \gzero\ to the identity function (so $\mathcal{G}_0(\mathcal{D})$ is original distribution $\mathcal{D}$) and \gone\ to a filter that adjusts the distribution to have a specified ratio over the desired attribute.
With respect to a binary property function, $f:\mathcal{X} \xrightarrow{} \bin$, $\mathcal{D}$ can be characterized using a generative probability density function:
\begin{align}
    \rho_{\mathcal{D}}(\textbf{x}) = \sum_{c \in \bin}p(c)\cdot p(\textbf{x}\;|\;c),
\end{align}
where $p(c)$ is a multinomial distribution representing the probabilities over the desired (binary) property function $f$ and its possible values $c$, and $p(\textbf{x}\,|\,c)$ is the generative conditional probability density function. Then, \gone$(\mathcal{D})$ can be expressed using the following probability density function, with a prior $\hat{p}$:
\begin{align}
    \rho_{G_1(\mathcal{D})}(\textbf{x}) = \sum_{c \in \bin}\hat{p}(c)\cdot p(\textbf{x}\;|\;c),\ \hat{p}(1) = \alpha,\ \hat{p}(0) = 1 - \alpha \label{eq:alpha}
\end{align}
where $\alpha$ is the probability of a randomly sampled point satisfying the property function $f$. Thus, a uniformly randomly sampled dataset from \gone$(\mathcal{D})$ would have an expected ratio of $\alpha$ of its members satisfying $f$. Additionally, we can modify \gzero\ with a similarly adjusted prior, enabling the adversary to distinguish between any two arbitrary ratios~\cite{zhang2021leakage}. In fact, our definition subsumes the one proposed by Chase \textit{et al.}~\cite{saeed} for the case of ratios over Boolean functions via the following instantiation:
\begin{align}
    D_{-},\ D_{+} & = \mathcal{G}_0(\mathcal{D}),\ \mathcal{G}_1(\mathcal{D}) \nonumber \\
    t_0,\ t_1 & = \alpha_0,\ \alpha_1,
\end{align}
along with setting $c=f(x)$ in Equation \ref{eq:alpha}.

Our definition, however, is not limited to describing proportional properties. For example, it can also be used to define the distributions over degrees for graph-based datasets, and infer properties of the underlying degree distributions. For this case, we represent graphs as samples from degree distributions: data with different degrees is sampled and then combined together in one graph. Since the adversary only cares about properties pertaining to the degrees of nodes (and not their attributes or other characteristics), it is safe to represent these samples purely in terms of degree distributions. This can then be used to target properties such as the mean-node degree of graphs, as we show in Section~\ref{sec:results}.

\section{Quantifying Leakage}\label{sec:quantifying}

Assessing the power of an attack is important for understanding it scientifically, and can also be of practical importance for both the victim and the adversary. 
Consider the most explored case in the literature---ratios of members satisfying a Boolean function. Intuition suggests that distributions with more different ratios (e.g., 0.2 and 0.9) would be easier to distinguish than more similar ones (e.g., 0.2 and 0.3), and most previous distributions inference results have focused on highly disparate distributions (often only showing meaningful distinguishing power when one of the ratios is at a 0.0 or 1.0 extreme).

Our framework enables us to quantify the amount of leakage observed in an attack by relating what an adversary is able to learn from a disclosed model to what they would learn from directly sampling examples from the training distribution.
As a setup, we provide the following lemma, proven in Appendix \ref{sec:appendix_theorem}, that gives an upper bound on the distinguishing accuracy (which we define as the probability of an adversary correctly inferring the underlying training distribution of a model) of any statistical test distinguishing between two distributions that differ in the proportion of records satisfying some Boolean property, using $n$ samples:
\begin{lemma}\label{lmm:distbound}
Given two Boolean-property proportional distributions \gzero($\mathcal{D}$), \gone($\mathcal{D}$) with proportion values $\alpha_0, \alpha_1$ derived from the same underlying distribution $\mathcal{D}$, the distinguishing accuracy between models trained on datasets of size $n$ from these distributions is at most
$$\frac{1}{2} + \frac{\min\left\{\sqrt{1 - \left(\frac{\min(\alpha_0, \alpha_1)}{\max(\alpha_0, \alpha_1)}\right)^n}, \sqrt{1 - \left(\frac{1 - \max(\alpha_0, \alpha_1)}{1 - \min(\alpha_0, \alpha_1)}\right)^n}\right\}}{2}.$$
\end{lemma}
First, we consider the most powerful possible adversary as one that can perfectly reconstruct training records from the model. The most that could be leaked to such an adversary is a perfect reconstruction of all the training records. 
Of course, we do not expect an adversary to reconstruct the training dataset fully, and an adversary can succeed in a high confidence distribution inference attack without being able to reconstruct any training records perfectly. Such a perspective is useful, though, for quantifying the power of an attack in a way that allows comparisons between attacks distinguishing distributions with different levels of variation. For some observed performance $\omega$ via an attack, we can compute the corresponding value of $n$ that would give an upper bound on accuracy as $\omega$. This value of $n$, which we term as \neff, thus quantifies the size of the dataset ``leaked" by the attack. In other words, it is equivalent to the adversary being able to draw \neff\ samples from the training distribution and executing an optimal distinguishing statistical test. The following theorem, proven in Appendix~\ref{app:theorem_followup}, shows how to compute \neff\ for an observed attack for the kind of distributions described above.

\begin{theorem} \label{thm:n_eff}
Given two Boolean-property proportional distributions \gzero($\mathcal{D}$), \gone($\mathcal{D}$) with proportion values $\alpha_0, \alpha_1$ derived from the same underlying distribution $\mathcal{D}$, and distinguishing accuracy $\omega$ using some attack,
$$
\neff\ =
\frac{\log (4\omega(1-\omega))}{\log (\max\left(\frac{\min(\alpha_0, \alpha_1)}{\max(\alpha_0, \alpha_1)}, \frac{1 - \max(\alpha_0, \alpha_1)}{1 - \min(\alpha_0, \alpha_1)}\right))}.
$$
\end{theorem}
A high value of \neff\ means the adversary is learning a lot about the underlying distribution, just using the given model. It helps put the attack's strength in perspective, given how similar the two distributions are. For instance, distinguishing between $\alpha_0 = 0.5$ and $\alpha_1 = 1.0$ with distinguishing accuracy $\omega=0.95$ corresponds to $\neff \approx 3$, whereas distinguishing between $\alpha_0 = 0.5$ and $\alpha_1 = 0.52$ with the same accuracy would correspond to $\neff \approx 42$. This aligns with intuition: the latter distribution is more similar and thus, should be ``harder'' for an attack to achieve the same kind of performance, and this notion is exactly what \neff\ aims to capture.

Note that this analysis is based on modeling an ``optimal attack'' where the adversary is able to directly sample records from the training distribution. This is just for deriving an expression for \neff, and not meant to assume any such attack. The expression above (and subsequent expressions for \neff) is a useful measure of any attack’s effectiveness that quantifies the leakage observed in the attack by relating it to the amount of information that would be leaked in an ``optimal attack'' where the adversary is just sampling training distribution records directly rather than inferring properties from a revealed model.

\shortsection{Regression over $\alpha$}
The case of distinguishing between ratios can be further extended to consider an adversary that wishes to directly predict the proportion value $\alpha$ for a given distribution. The following theorem, proven in Appendix~\ref{app:regression}, shows how to compute \neff\ for an observed attack with square error $\omega$.

\begin{theorem} \label{thm:regress_neff}
Given a Boolean-property proportional distribution with proportion value $\alpha$, and square error $\omega$ observed using some attack,
$$
\neff\ = \frac{\alpha(1-\alpha)}{\omega}.
$$
\end{theorem}
This result extends our notion of \neff\ to adversaries that directly infer the underlying ratio $\alpha$, a more realistic adversary goal that we also explore in our experiments.

\shortsection{Graphs as Distributions of Natural Numbers}
Similar to the ratio case, our framework enables us to compute \neff\ when working with distributions of natural numbers. For the purpose of distinguishing between graph distributions, this notion of `distribution of numbers' can be extended to graphs by studying their degree distributions. As a setup, we provide the following lemma, proven in Appendix \ref{sec:appendix_degree_theorem}, that gives an upper bound on the distinguishing accuracy of any statistical test distinguishing between two distributions following Zipf's law, using $n$ samples:

\begin{lemma}\label{lmm:distbound_degree}
Given two distributions of natural numbers, \gzero($\mathcal{D}$) and \gone($\mathcal{D}$) that follow Zipf's law, with $N_0$ and $N_1$ elements (without loss of generality assume $N_0 \leq N_1$) and parameters $s_0, s_1$
respectively, the distinguishing accuracy between models trained on graphs with $n$ nodes from these distributions is at most:
$$\frac{1}{2} + \frac{\sqrt{1-\left(\frac{H_{N_0, s_0}}{H_{N_1, s_1}}N_0^{(s_0-s_1)\mathbb{I}[s_1 > s_0]}\right)^n}}{2}.$$
\end{lemma}
Here, $H_{n, s}$ is the $n^{th}$ generalized Harmonic number of order $s$,
$H_{n, s} = \sum_{k=1}^N k^{-s}$.  Together, these two parameters are related to the expected mean of the distribution $\alpha_b$ (mean node-degree, in the case of degree distributions) as:
\begin{align}
    \alpha_b = \frac{H_{N_b, s_b-1}}{H_{N_b, s_b}} \label{eq:mean_harmonic_relation}
\end{align}

Similar to the case of different Boolean-property ratios distributions, we can compute \neff\ for a given attack and its observed performance (see Appendix~\ref{app:theorem_degree_followup} for the proof):
\begin{theorem} \label{thm:n_eff_degree}
Given two distributions of natural numbers
distributions \gzero($\mathcal{D}$), \gone($\mathcal{D}$) that follow Zipf's law, with $N_0, N_1$ elements (without loss of generality assume $N_0 \leq N_1$) and parameters $s_0, s_1$ respectively, and observed distinguishing accuracy $\omega$, 
$$
\neff = \frac{\log(4\omega(1-\omega))}{\log\left(\frac{H_{N_0,s_0}}{ H_{N_1, s_1}}\right) + (s_0 - s_1)\mathbb{I}[s_1 > s_0]\log(N_0)}.
$$
\end{theorem}
Compared to the ratio distinguishing attacks, attacks on the \graph\ dataset are much more successful: reaching near-perfect distinguishing accuracies as well as the highest \neff\ numbers (Table~\ref{table:datasets_desc}).

\section{Attacks}
\label{sec:attacks}

In our experiments to evaluate the risks of property inference attacks, we use two simple black-box attacks and the state-of-the-art white-box meta-classifier attack based on Permutation-Invariant Networks~\cite{ganju2018property}. We extend the meta-classifier attack to support convolutional neural networks (Section~\ref{sec:attack_conv}). These attacks do not assume anything about the underlying distributions they try to differentiate other than the availability of the underlying distribution and knowledge of the public \gzero\ and \gone\ transformers. For the white-box attacks, full access to the trained model is assumed; for the black-box attacks, the ability to sample data from distributions \gzero\ and \gone is assumed.

\subsection{Black-Box Attacks}

The black-box attacks assume the adversary has access to representative data for the candidate distributions, but only has API access to the target model which outputs its prediction (just the label) for a submitted input. Although access to model parameters is unavailable in such an approach, its model-agnostic nature allows adversaries to launch attacks on any target model.

\subsubsection{\ltest}
A simple algorithm $\mathcal{H}$ is to test the accuracy of the model on datasets from the two candidate distributions, and conclude that the training distribution is closest to whichever test dataset the model performs better on. For data samples $S_{b \in \{0, 1\}} \sim \mathcal{G}_b(\mathcal{D})$:
\begin{align}
    \hat{b} = \mathbb{I}[\acc(M, S_0) < \acc(M, S_1)], \label{eq:loss_test}
\end{align}
where $\acc(M, S)$ is the accuracy of model $M$ on some sample of data $S$, and  $\mathbb{I}$ is the indicator function. Intuitively, a model would have higher accuracy on data sampled from the training distribution, compared to another distribution. This method does not require the adversary to train models, but only to have access to suitable test distributions, and the ability to submit samples from those distributions to the target model. The data held by the adversary here is not overlapping with the data used by the victim to train its models, thus ruling out any potential for leakage via shared data. Although we use the accuracy in Equation~\ref{eq:loss_test}, the adversary can use any other metric that captures model performance, like the loss used to train these models.
\begin{figure*}[tbh]
\centering
\includegraphics[width=0.95\textwidth]{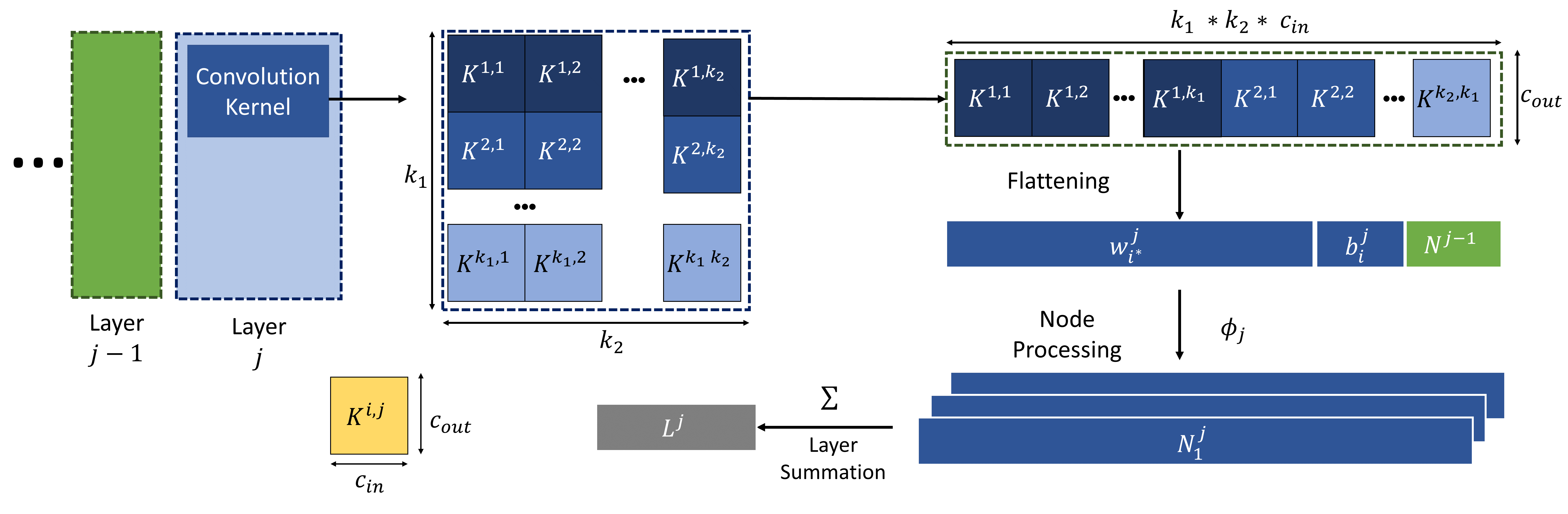}
\caption{Transforming a $k_1 \times k_2$ kernel matrix $K$ (with input channels $c_{in}$ and output channels $c_{out}$) into a 2-dimensional weight matrix for compatibility with the Permutation Invariant Network architecture. The node-processing functions $\phi_i$ and the rest of the pipeline are identical to the Permutation Invariant Network described in Section~\ref{sec:meta_classifiers}.}
\label{fig:pin_conv}
\end{figure*}
\subsubsection{\ttest}
The \ltest\ assumption may not hold for some pairs of distributions if one distribution is inherently easier to classify than the other (as we observe in our experiments in Section~\ref{sec:results}).
To account for this, we consider an attack where the adversary trains and uses a small (balanced) sample of models from each distribution to identify which of $S_0$ or $S_1$
maximizes the performance gap between its models.

\begin{align}
    \gamma_{c \in \{0, 1\}} & = \sum_{i; y_i = 0}{\acc(M^i, S_c)} - \sum_{i; y_i = 1}{\acc(M^i, S_c)} \nonumber \\
    k & = \mathbb{I}[\abs{\gamma_0} < \abs{\gamma_1}], \label{eq:ttest_1}
\end{align}
\noindent
where $y_i \in \{0,1\}$ denotes that the model $M^i$ is trained on a dataset sampled from $\mathcal{G}_{\{0,1\}}(\mathcal{D})$ respectively.  After identifying $k_{\in \{0, 1\}}$, the adversary derives a threshold $\lambda$ to maximize accuracy distinguishing between models trained on datasets from the two distributions (using a simple linear search). Assuming $\gamma_k$ is positive,
\begin{align}
    \delta(\Lambda) & = \sum_{i;y_i=0}\mathbb{I}[\acc(M^i, S_k) \geq \Lambda] + \sum_{i;y_i=1}\mathbb{I}[\acc(M^i, S_k) < \Lambda] \nonumber \\
    \lambda & = \argmax_{\Lambda} \delta(\Lambda). \label{eq:threshold}
\end{align}
The adversary then predicts $\hat{b}=\mathbb{I}[\acc(M, S_k) \geq \lambda]$ (or with a $<$ inequality when $\gamma_k$ is negative). Thus, the adversary uses a sample of locally-trained models to derive a classification rule based on model accuracy, which it then uses to infer the training distribution of the target model. For the same reasons as \ltest\ the data used by the adversary here, for both training its local set of models and computing the threshold), is non-overlapping with data used by the victim. Similar to the \ltest, we can use any other metric instead of accuracies in the \ttest\   (Equations~\ref{eq:ttest_1}, \ref{eq:threshold}) that capture model performance, like the loss used to train these models.

\subsection{White-Box Attacks}\label{sec:whiteboxattacks}

The white-box attacks assume the adversary has direct access to the trained model, including its parameters, and has access to a large enough amount of representative data to train local models on the candidate distributions. Although access to the model itself is a stronger assumption than the black-box approach, it can potentially reveal much more information about the training distribution via model parameters.

\subsubsection{Meta-Classifiers} 
\label{sec:meta_classifiers}
The state-of-the-art property inference attack is Ganju et al.'s attack using Permutation-Invariant Networks as meta-classifiers~\cite{ganju2018property}. The meta-classifiers take as input model parameters (weights, bias) and predict which distribution was used to sample training data for the model. This architecture is designed to be invariant to different neuron orderings inside neural network layers, which it achieves by utilizing the DeepSets~\cite{zaheer2017deep} architecture. Neuron-ordering invariance is achieved via a set of transforming functions, $\phi_i$ (defined for each layer $i$), over each row of the layer weight matrix. The outputs of these functions are then summed to create a layer representation $L^ i$, achieving invariance to the ordering of the neurons within each layer via the summation function.
Since the meta-classifier is itself a classifier that requires many models (800 per distribution Ganju et al.'s work~\cite{ganju2018property}) trained on the two distributions for training, this attack is only feasible for adversaries with access to sufficient data from both training distributions and considerable computational resources.

\subsubsection{Targeting Convolutional Neural Networks}
\label{sec:attack_conv}

The Permutation-Invariant Network only supports linear layers in a feed-forward architecture; previous work only considered two or three layer MLPs on small datasets~\cite{ganju2018property}, or single-layer recurrent neural networks~\cite{zhang2021leakage}. Applying the same architecture on top of convolutional layers requires some adaptations since kernel matrices are four-dimensional. While there is permutation invariance within channels, the kernel itself is sensitive to permutations by the nature of how the convolution operation works. We extend property inference attacks to convolutional neural networks and demonstrate their effectiveness on deep-learning models with up to eight layers, trained from scratch.

Figure~\ref{fig:pin_conv} illustrates our method. Let $K$ be a kernel of size $(k_1, k_2)$ associated with some convolutional layer, with input and output channel dimensions $c_{in}$ and $c_{out}$ respectively. While designing the architecture to capture invariance, it is important to remember that positional information in the kernel matters, unlike neurons in a linear layer. Thus, any attempt to capture invariance should be limited to the mapping between input and output channels of a convolutional kernel. With this in mind, we flatten the kernel of size $(k_1, k_2, c_{in}, c_{out})$ such that the resulting matrix is of size $(k_1 \times k_2 \times c_{in}, c_{out})$. Concatenating along the input channel dimension helps preserve location-specific information learned by the kernel while capturing permutation invariance across the output channels themselves.

This two-dimensional matrix is then processed in the same way as the linear layers are in Permutation-Invariant Networks (using the same notation as Section~\ref{sec:meta_classifiers}), applying function $\phi_i$ and summing to capture invariance while concatenating feature representations from prior layers. Similar to the original architecture, the bias component can be concatenated to the kernel matrix itself. Since this feature extraction process on a convolutional layer also produces a layer representation, it can be easily incorporated into the existing architecture to work on models with a combination of convolutional and linear layers.
\section{Experiments}
\label{sec:experiments}

To better understand the risks of distribution inference, we execute distribution inference attacks to measure their ability to distinguish distributions with varying disparity on tabular, image, and graph datasets. Although it is unknown how close these attacks are to the best possible distribution inference attacks, they help demonstrate an empirical lower bound on adversarial capabilities and can be helpful in estimating general trends. Code
for reproducing our experiments is available at:
\url{https://github.com/iamgroot42/FormEstDistRisks}.

\subsection{Datasets}
\label{sec:datasets}
\renewcommand{\bdv}{\mathcal{T}}
We report on experiments using five datasets, summarized in Table~\ref{table:datasets_desc}. We construct non-overlapping data splits between the simulated adversary, $\adv$, and model trainer $\bdv$.
These non-overlapping splits help better capture a realistic scenario where the adversary has access to training data from the distribution $\mathcal{D}$ but is unlikely to have any of the model trainer's training data (which is considered private). Both parties then modify their dataset to emulate a distribution property, and then sample training datasets from these adjusted distributions to train and evaluate their models. This sampling, along with the disjoint data splits between $\adv$ and $\bdv$, helps ensure that any distinguishing power we observe is actually distribution inference, rather than inadvertent dataset inference.

\begin{table}
\small
\centering
\begin{tabular}{c c c c c c c c} 
 \toprule
 \multirow{2}{*}{\textbf{Dataset}} & \multirow{2}{*}{\textbf{Task}} & \multirow{2}{*}{\textbf{Property}} & \multicolumn{2}{c}{\textbf{Size}} &
 \multicolumn{2}{c}{\textbf{Binary}} & \textbf{Regression} \\ 
  & & & Adversary & Victim & \neff & $\|\alpha_0 - \alpha_1\|_{0.75}$ & \neff \\
 \midrule
\multirow{2}{*}{\census} & \multirow{2}{*}{Income prediction} & Ratio female & 3200 & 3200 & 0.2 & 0.5 & 8.8 \\
 &  & Ratio white & 1800 & 1800 & 0.1 & 0.6 & 6.0 \\
 \midrule
 \multirow{2}{*}{\celeba} & Smile identification & Ratio female & 22,000 & 36,000 & 0.3 & 0.5 & 10.6\\
  & Gender prediction & Ratio young & 6700 & 18,800 & 0.2 & 0.6 & 5.1\\
 \midrule
 \boneage & Age prediction & Ratio female & 1800 & 3600 & 6.2 & 0.2 & 269.4\\
 \midrule
 \graph & \multirow{2}{*}{Node classification} & Node degree & 40,000 & 97,000 & 30.6 & - & - \\
 \botnet & & Clustering & 135,216 & 135,523 & - & - & - \\
 \bottomrule
\end{tabular}
\caption{Descriptions of datasets, along with the effectiveness of inference attacks (best of all \ltest, \ttest, and meta-classifier for binary classification, and meta-classifier for regression) while varying ratios of distributions. $\|\alpha_0 - \alpha_1\|_{0.75}$ is the minimum difference in ratios observed that has at least 75\% average accuracy. For binary classification, \neff\ is the median effective $n$ value based on Theorem~\ref{thm:n_eff} using the maximum distinguishing accuracies across all experiments and pairs of property ratios (degrees in the case of \graph) without outliers. For regression, \neff\ is the mean effective $n$ value based on Theorem~\ref{thm:regress_neff} across all ratios excluding 0 and 1. Size for \graph\ refers to number of nodes, and the average number of nodes for \botnet.}
\label{table:datasets_desc}
\end{table}

Our experimental datasets were selected to incorporate common benchmarks (\census, \celeba) to enable comparisons with previous work, new datasets to assess more realistic property inference threats (\boneage), as well as applying our definitions beyond ratio-based properties on graphs (mean node-degree on \graph, clustering coefficient on \botnet). As noted by Zhang et al.~\cite{zhang2021leakage}, the target properties for a distribution inference attack can be either related to or independent of the task and can be either explicit features of the input data or latent features. For instance, attributes varied for \census\ are feature-based properties since these attributes are directly used as features for the models trained on them. On the other hand, an attribute like the age of a person is unrelated to detecting smiles and is a latent property that is not directly encoded as an input feature in the training data (but is available for our \celeba\ experiments from provided metadata).

\textbf{\census}\ \cite{bay2000uci} consists of several categorical and numerical attributes like age, race, education level to predict whether an individual's annual income exceeds \$50K. We focus on the ratios of whites (race) and females (sex) as properties and use a three-layer feed-forward network as the model architecture.

\textbf{\boneage}\ \cite{halabi2019rsna} contains X-Ray images of hands, with the task being predicting the patient's age in months. We convert the task to binary classification based on an age threshold and use a pre-trained DenseNet~\cite{huang2017densely} model for feature extraction, followed by a two-layer network for classification. We focus on the ratios of the females (available as metadata, but implicit in the examples) as properties.

\textbf{\celeba}\ \cite{liu2018large} contains face images of celebrities, with multiple images per person. Each image is annotated with forty attributes such as gender, sunglasses, and facial hair. We use two different tasks, smile detection and gender prediction, and train convolutional neural networks from scratch for this dataset. We focus on the ratios of the females (smile-detection task) and old people (gender-prediction task) as properties. These attributes are provided as meta-data in the dataset. 
We create a network architecture with five convolutional layers and pooling layers followed by three linear layers, which is the smallest one we could find with reasonable task accuracy.

The \textbf{\graph}~\cite{wang2020microsoft} dataset is a directed graph, representing citations between computer science arXiv papers. The task is to predict the subject area categories for unlabeled papers. We infer the mean node-degree property of the graph. We use a four-layer Graph Convolutional Network~\cite{kipf2017semi}.

\textbf{\botnet}\ \cite{zhou2020auto} contains botnets with the Chord~\cite{stoica2001scalable} topology artificially overlaid on top of background network traffic from CAIDA~\cite{caida2018caida}. The dataset contains multiple graphs, with the task of detecting bot nodes in the graphs. We focus on inferring whether the underlying graphs (onto which we overlay botnets) have average clustering coefficients within a specific range. Following the model architecture proposed in Zhou et al.~\cite{zhou2020auto}, we implement a Graph Convolutional architecture.

\subsection{Attack Details}
\label{sec:attack_details}

We perform each experiment ten times and report mean values with standard deviation in all of our experiments. Since the \ltest\ uses a fixed test set per experiment, its results show no variation. For each dataset, we train 1000 victim models per distribution.

\shortsection{Loss Test} The adversary uses its test data to sample the two test sets $S_0$ and $S_1$. Since we use the same test data in evaluations, we turn off sampling while generating data with desired properties for this setting.

\shortsection{Threshold Loss}
The adversary trains 50 models per distribution on its data split.

\shortsection{Meta-Classifier} We used Permutation Invariant Networks as our meta-classifier architecture~\cite{ganju2018property}. The simulated adversary produces 800 models per distribution using its split of data to train the meta-classifier. For the case of \celeba, we use our extension of the Permutation Invariant Network that is compatible with convolutional layers (Section~\ref{sec:attack_conv}). Following experimental designs from prior works, we were able to achieve the accuracies that the authors reported (Section~\ref{sec:distinguishing_ratios}).
However, using our experimental design leads to significantly lower distinguishing performance. Steps like ensuring no overlap in victim/adversary training data, randomly sampled datasets for \gzero$(\mathcal{D})$ and \gone$(\mathcal{D})$, and ensuring the same dataset size are necessary to avoid the risk that the meta-classifier is identifying something different about the distributions other than the claimed property.
We think these steps are important for realistic experiments, so report the distinguishing accuracies based on this experimental design, even if they are lower for the same attacks than the results reported for the same tasks in previous work.

\subsection{Binary Properties}
\label{sec:results}

\begin{figure*}
\centering
\begin{subfigure}[b]{0.327\textwidth}
    \centering
    \includegraphics[width=\textwidth]{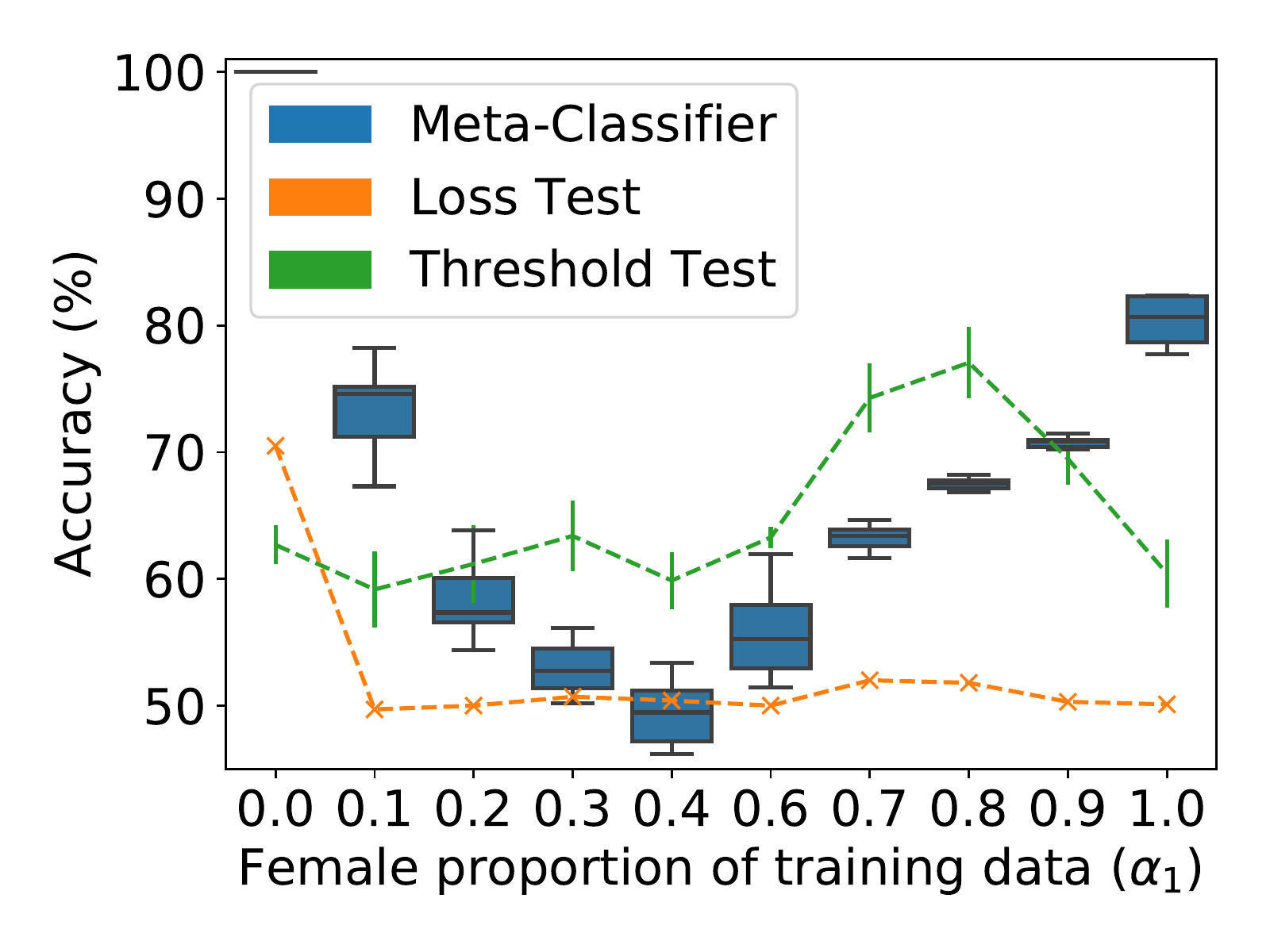}
    \caption{\census}
    \label{fig:distinguishing_female_proportion_census}
\end{subfigure}
\hfill
\begin{subfigure}[b]{0.327\textwidth}
    \centering
    \includegraphics[width=\textwidth]{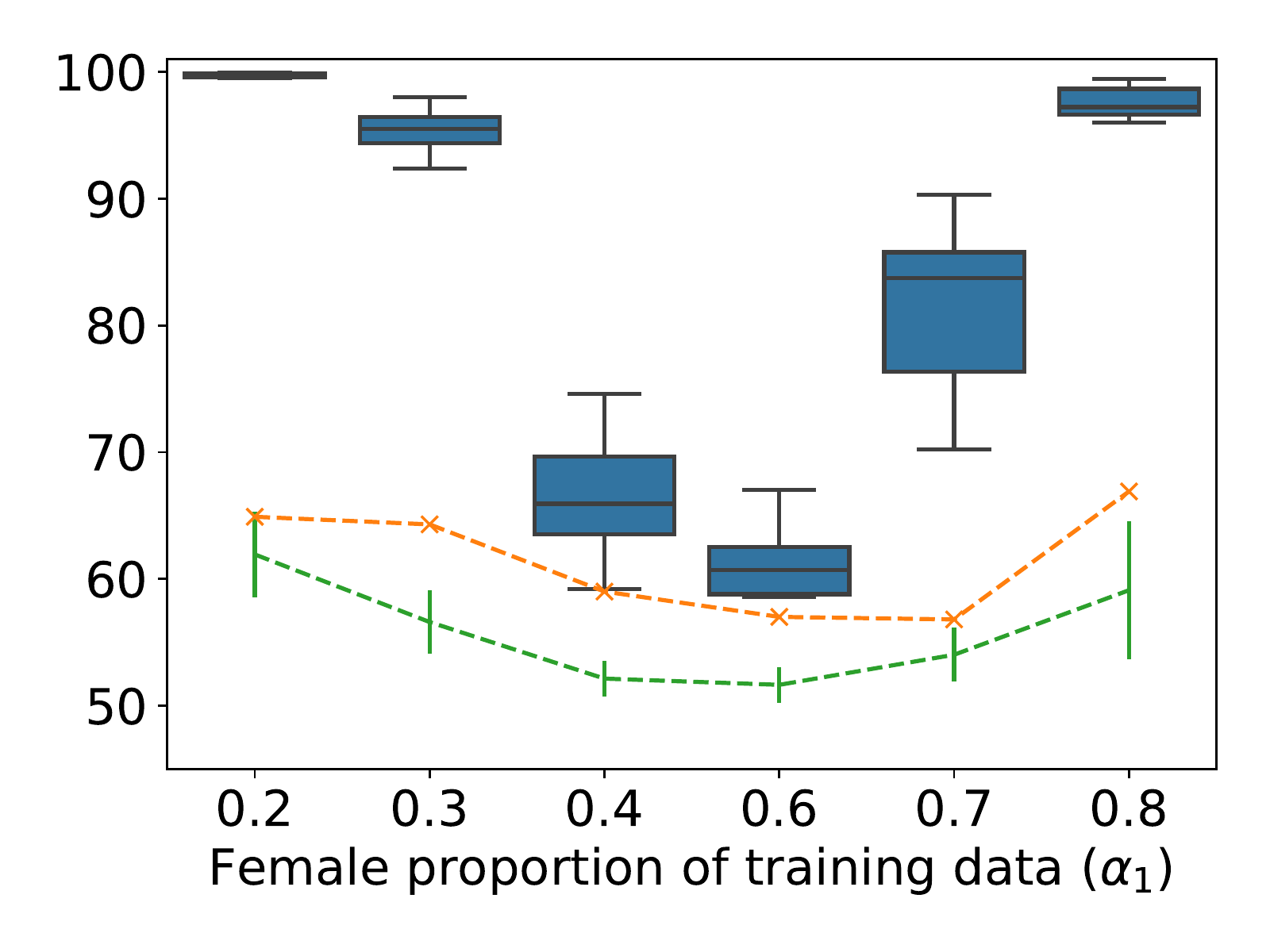}
    \caption{\boneage}
    \label{fig:distinguishing_female_proportion_boneage}
\end{subfigure}
\hfill
\begin{subfigure}[b]{0.327\textwidth}
    \centering
    \includegraphics[width=\textwidth]{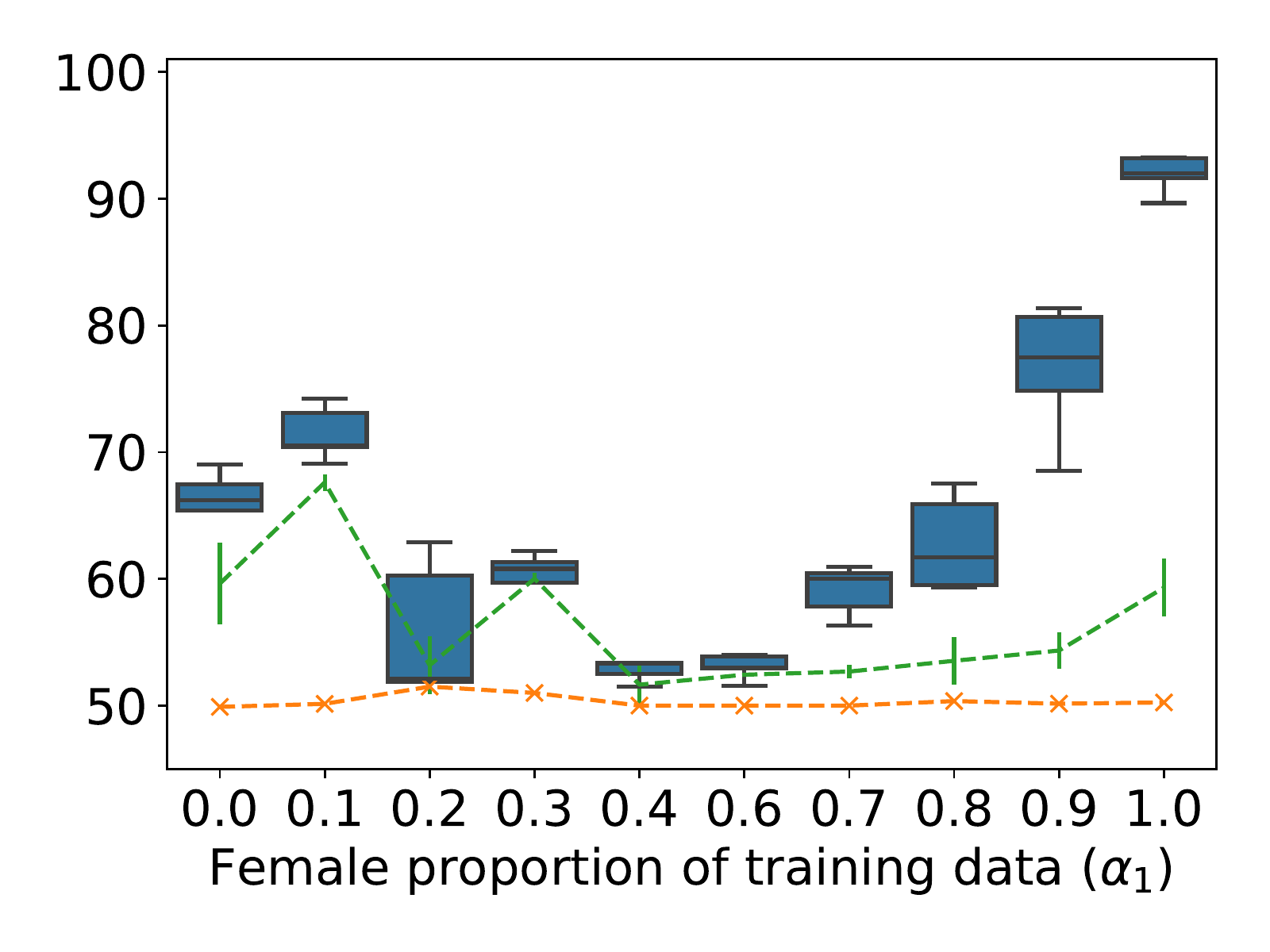}
    \caption{\celeba}
    \label{fig:distinguishing_female_proportion_celeba}
\end{subfigure}
\caption{Classification accuracy for distinguishing proportion of females in training data for (\protect{\subref{fig:distinguishing_female_proportion_census}}) \census, (\protect{\subref{fig:distinguishing_female_proportion_boneage}}) \boneage, and (\protect{\subref{fig:distinguishing_female_proportion_celeba}}) \celeba. 
The \boneage\ dataset does not include ratios below 0.2 or above 0.8, since sampling the original dataset for that ratios produces datasets that are too small to train models with meaningful performance. Performance of the attacks increases as the distributions diverge ($\alpha_1$ moves away from 0.5), but is not symmetric.
}
\label{fig:distinguishing_female_proportion}
\end{figure*}

We fix \gzero\ and try the attacks on a range of \gone\ distributions for the first set of experiments. In Section~\ref{sec:finegrained}, we report on experiments varying both distributions. Most prior works on distribution inference
use arbitrary ratios, like distinguishing between 42\% and 59\% males~\cite{ganju2018property} while executing property inference attacks. Only recently have works started transitioning to more controlled experimental settings, like comparable dataset sizes for the victim and adversary and non-overlapping sampling of data~\cite{zhou2021property}. While having one of the ratios corresponding to the estimate of the underlying data distribution is justified, fixing the other arbitrarily makes it hard to understand the adversary's capabilities---we want to understand how dissimilar the distributions must be in order to be distinguishable. Additionally, the lack of using the same ratios in experiments while looking at different properties makes it harder to compare how much information about one property is leaked by models compared to the other. Analyzing such trends is important for understanding how much of these properties are leaked across different configurations (explicit attribute, latent property) and assess the adversary's capabilities under different scenarios (black-box access, white-box access).

Experiments with binary classification for properties provide a simple goal that can provide useful insights into the effectiveness of distribution inference attacks, but most realistic attacks would not be based on distinguishing between two known distributions. In Section~\ref{sec:regression_results}, we consider attacks that can infer the underlying ratio without any prior assumptions about distinguishing particular distributions.


\subsubsection{Distinguishing Imbalanced Ratios}
\label{sec:distinguishing_ratios}

Since the original ratios for the targeted property may be unbalanced in the dataset (Section~\ref{sec:existing_defns}), for these experiments set the fixed \gzero\ to a balanced ($\alpha_0 = 0.5$) ratio for the chosen attribute for the Census, \celeba, and \boneage~datasets. Then, we vary $\alpha_1$ (Equation~\ref{eq:alpha}) to evaluate inference risks and understand how well an adversary could distinguish between models trained using distributions with different proportions of the targeted property. 
\begin{figure}
\centering
\begin{subfigure}[b]{0.48\textwidth}
    \centering
    \includegraphics[width=\textwidth]{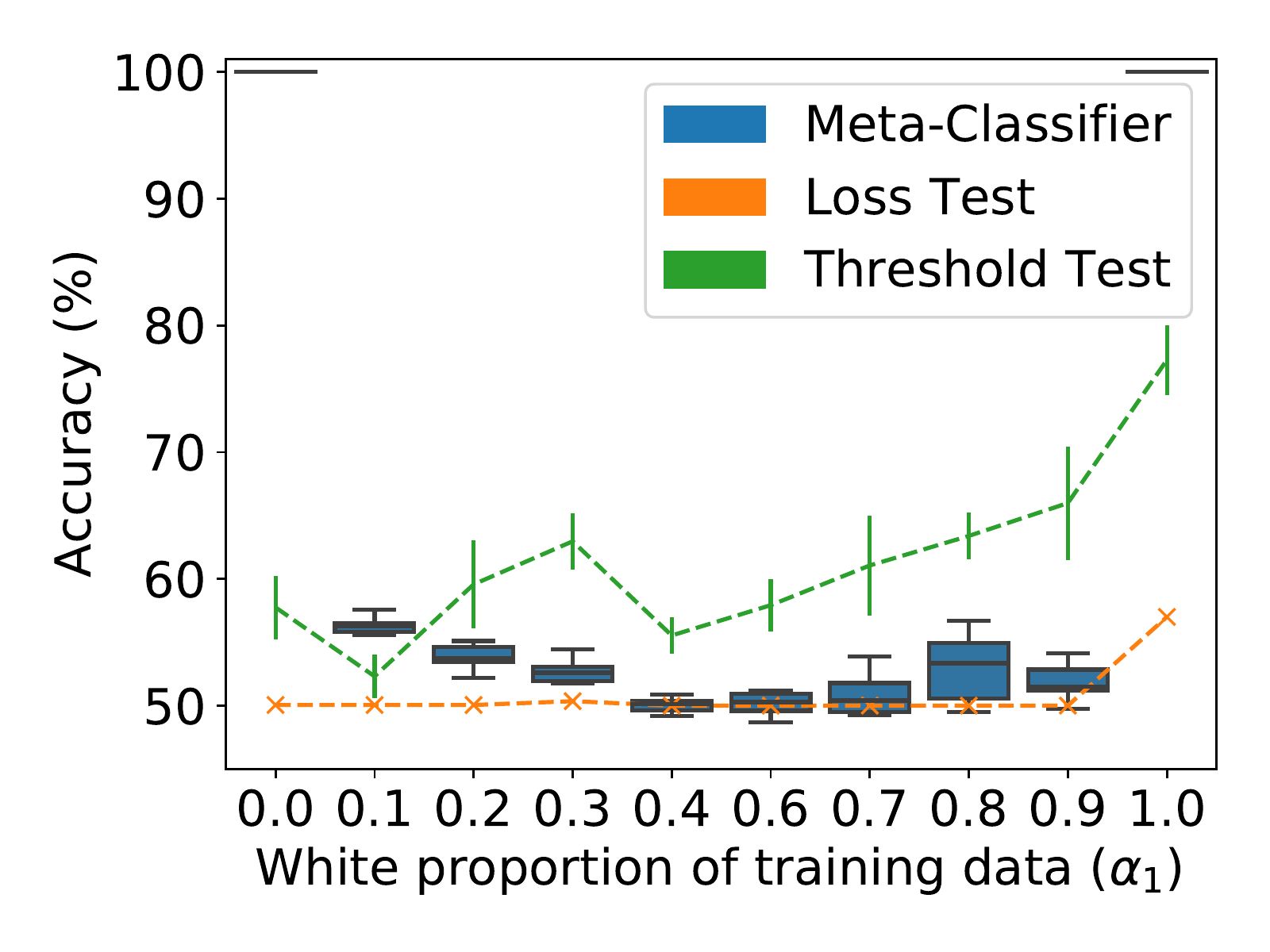}
    \caption{\census, race}
    \label{fig:distinguishing_other_proportions_census}
\end{subfigure}
\hfill
\begin{subfigure}[b]{0.48\textwidth}
    \centering
    \includegraphics[width=\textwidth]{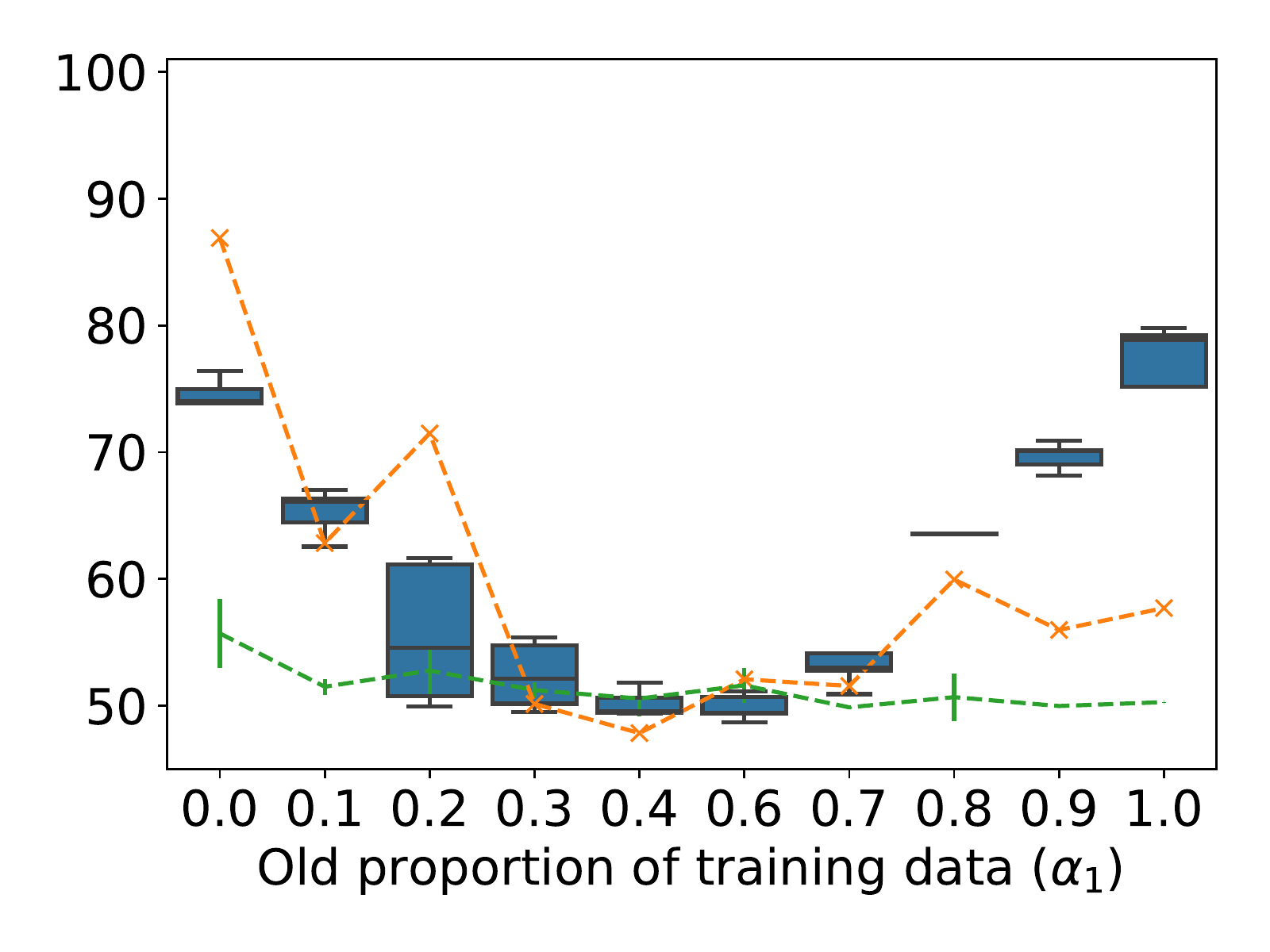}
    \caption{\celeba, age}
    \label{fig:distinguishing_other_proportions_celeba}
\end{subfigure}
\caption{Distinguishing accuracy for proportion of (\protect\subref{fig:distinguishing_other_proportions_census}) whites in training data for the \census\ and (\protect\subref{fig:distinguishing_other_proportions_celeba}) old people in the \celeba. The black-box attacks approach the performance of the white-box meta-classifier attacks for some distributions (especially the \ttest\ and the \census\ (race) task), further motivating the importance of considering computationally cheaper black-box attacks.}
\label{fig:distinguishing_other_proportions}
\end{figure}

\shortsection{\census}
We summarize the accuracies for the three attack methods across varying proportions of females in Figure~\ref{fig:distinguishing_female_proportion_census} and whites in Figure~\ref{fig:distinguishing_other_proportions_census}. The \ltest\ performs only marginally better than random guessing in most cases for females, yielding \neff\ values in the range $[0, 0.03]$, essentially close to random guessing. On the other hand, the \ttest\ and meta-classifiers are able to achieve non-trivial distinguishing accuracies for the female proportion, with \neff\ values in ranges $[0.1, 1.2]$ and $[0.02, 6.5]$ respectively, with a similar median \neff\ value of 0.33, showing how the two are not very far apart in effectiveness. Leakage increases as the distributions become more disparate, with near-perfect distinguishing accuracy for the extreme case of $\alpha_1=0$.
For race, none of the attacks detect anything (\neff$\approx0$) apart from the surprising results on \ttest, which performs asymmetrically well, approaching 80\% accuracy for mostly-white distributions.

\shortersection{Comparison with previous results}
Ganju et al.~\cite{ganju2018property}
applied their meta-classifier method on two properties on this dataset: 38\% vs.\ 65\% women (case A), and 0\% vs.\ 87\% whites (case B). For case B, the \ltest\ performs as well as random guessing (50.1\%), while the \ttest\  ($92.4\pm2.6$\%) approaches meta-classifier performance ($99.9 \pm 0.1$\%), achieving a high (compared to $\alpha=0.5$ experiments) \neff$\approx11$. For case A, the \ttest\ ($62.7 \pm 2.0$\%) outperforms meta-classifiers ($62.1 \pm 1.7$\%) while achieving \neff$\approx0.03$, barely better than random guessing , and the \ltest\ method fails (50\%). Ganju et al.\ report 97\% accuracy for case A and 100\% for case B. We were able to closely reproduce these results in their setting which includes overlapping data between victim and adversary and does not ensure changes in ratios do not affect dataset size or class imbalance. In the more realistic experimental design in which we ensure there is no victim/adversary overlap and maintain the label ratios and same dataset sizes (Section~\ref{sec:attack_details}), the accuracies are much lower---for example, in case A the distinguishing accuracy is $97\%$ using their experimental design but drops to $62\%$ when more carefully prepared datasets are used in our design. These results suggest that although the distinguishing accuracy is high between the two distributions in the tests in their setting, the attacks are not actually inferring the intended property but are predicting the distribution based on other differences between the datasets. 

\shortsection{\boneage}
Distinguishing accuracies for the female proportion on the RSNA Bone Age dataset are plotted in Figure~\ref{fig:distinguishing_female_proportion_boneage}. The simple \ltest\ performs nearly as well as the \ttest\, leaking a median of \neff$\approx0.2$ and \neff$\approx0.1$ respectively. The meta-classifier attack, on the other hand, has a much higher leakage of \neff$\approx6$.

\begin{figure}[bt]
\centering
\begin{subfigure}[b]{0.48\textwidth}
         \centering
         \includegraphics[width=\textwidth]{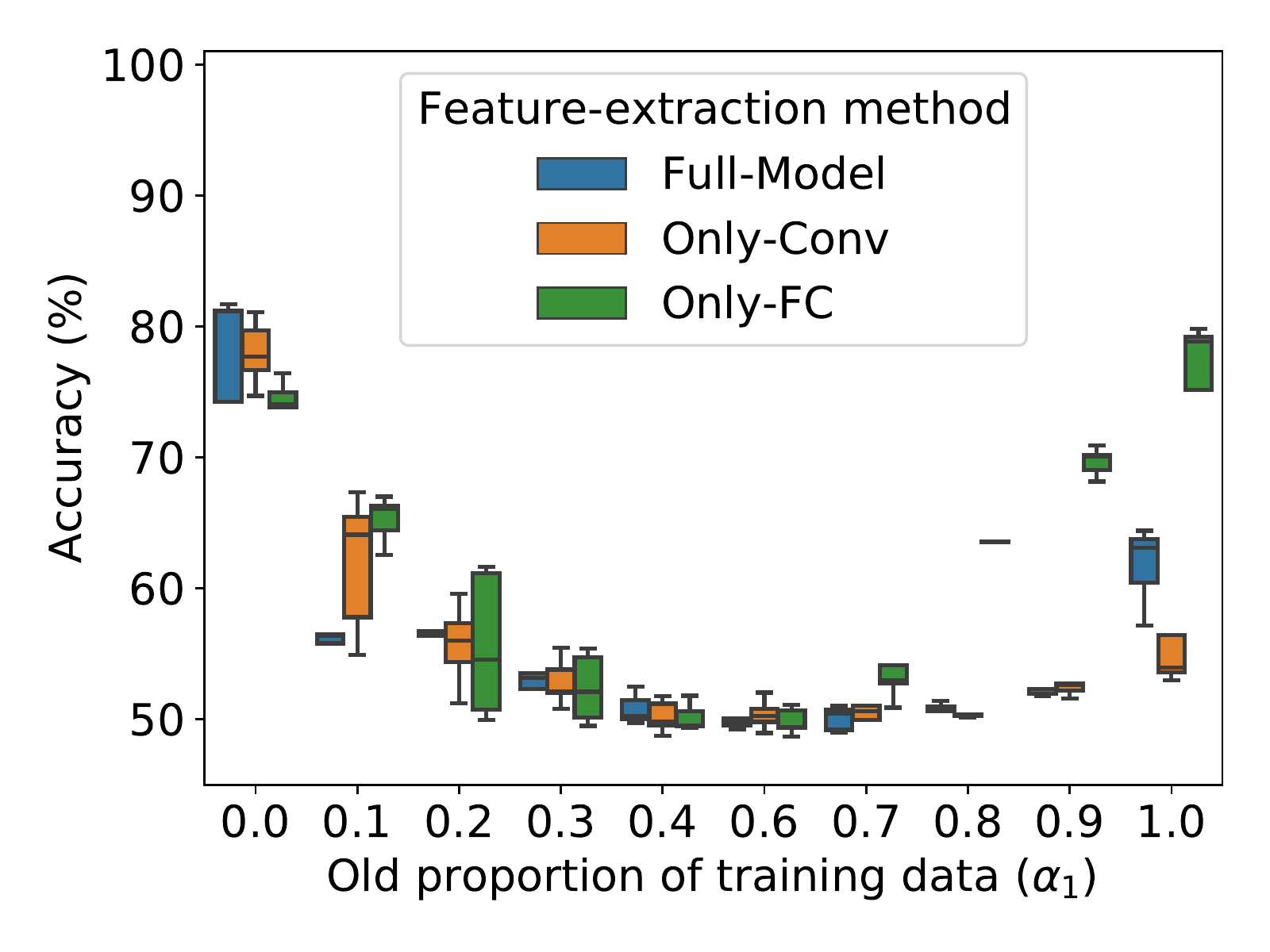}
         \caption{\celeba, age}
         \label{fig:celeba_meta_boxplots_old}
     \end{subfigure}
     \hfill
     \begin{subfigure}[b]{0.48\textwidth}
         \centering
         \includegraphics[width=\textwidth]{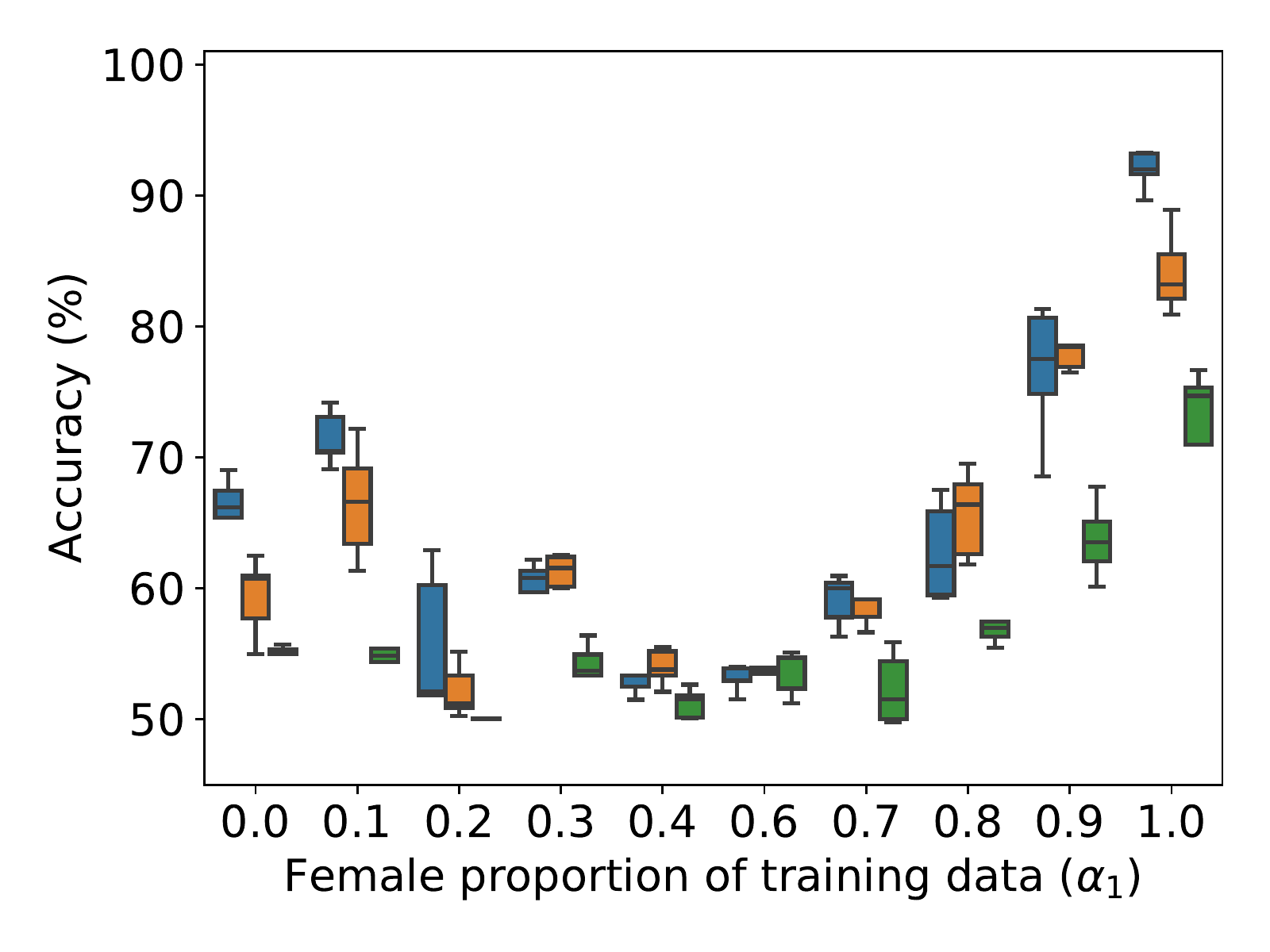}
         \caption{\celeba, females}
         \label{fig:celeba_meta_boxplots_females}
\end{subfigure}
\caption{Distinguishing accuracy for meta-classifiers for proportion of (\protect\subref{fig:celeba_meta_boxplots_old}) old people and (\protect\subref{fig:celeba_meta_boxplots_females}) females in the training data for \celeba. Using all the layers' parameters is not necessarily helpful and can lead to lower performance (e.g., \celeba, females).}
\label{fig:celeba_meta_boxplots}
\end{figure}

\shortsection{\celeba}
Figure~\ref{fig:distinguishing_female_proportion_celeba} shows the distinguishing accuracy for the \celeba\ data on proportion of females, and Figure~\ref{fig:distinguishing_other_proportions_celeba} for the proportion of examples marked as ``old''. The \ttest\ performs much worse compared to \census\ and \boneage, with the median \neff$<0.05$, compared to $\approx0.7$ using meta-classifiers.
Figure~\ref{fig:celeba_meta_boxplots} shows meta-classifier prediction accuracy for three different representations of the shadow models used to train the meta-classifier: using parameters from only linear layers, only convolutional layers, and all layers of the models. While inferring the ratio of old people (Figure~\ref{fig:celeba_meta_boxplots_old}), including just the fully-connected layers works best, yielding \neff$\approx0.12$ and not too far off from using all layers or just the convolutional layers (\neff$\approx0.07$). For the case of sex ratios (Figure~\ref{fig:celeba_meta_boxplots_females}), using the full model helps extract more information (\neff$\approx0.32$) than either of the convolutional (\neff$\approx0.24$) or linear (\neff$\approx0.05$) layers.
These trends suggest the likelihood of some layers' parameters capturing specific property-related information better than the others. Linear layers are more helpful for ratios of old people whereas for ratios of females, convolutional layers are significantly better.

\begin{figure*}[!bt]
\centering
\begin{subfigure}[b]{0.49\textwidth}
    \centering
\includegraphics[width=1.0\textwidth]{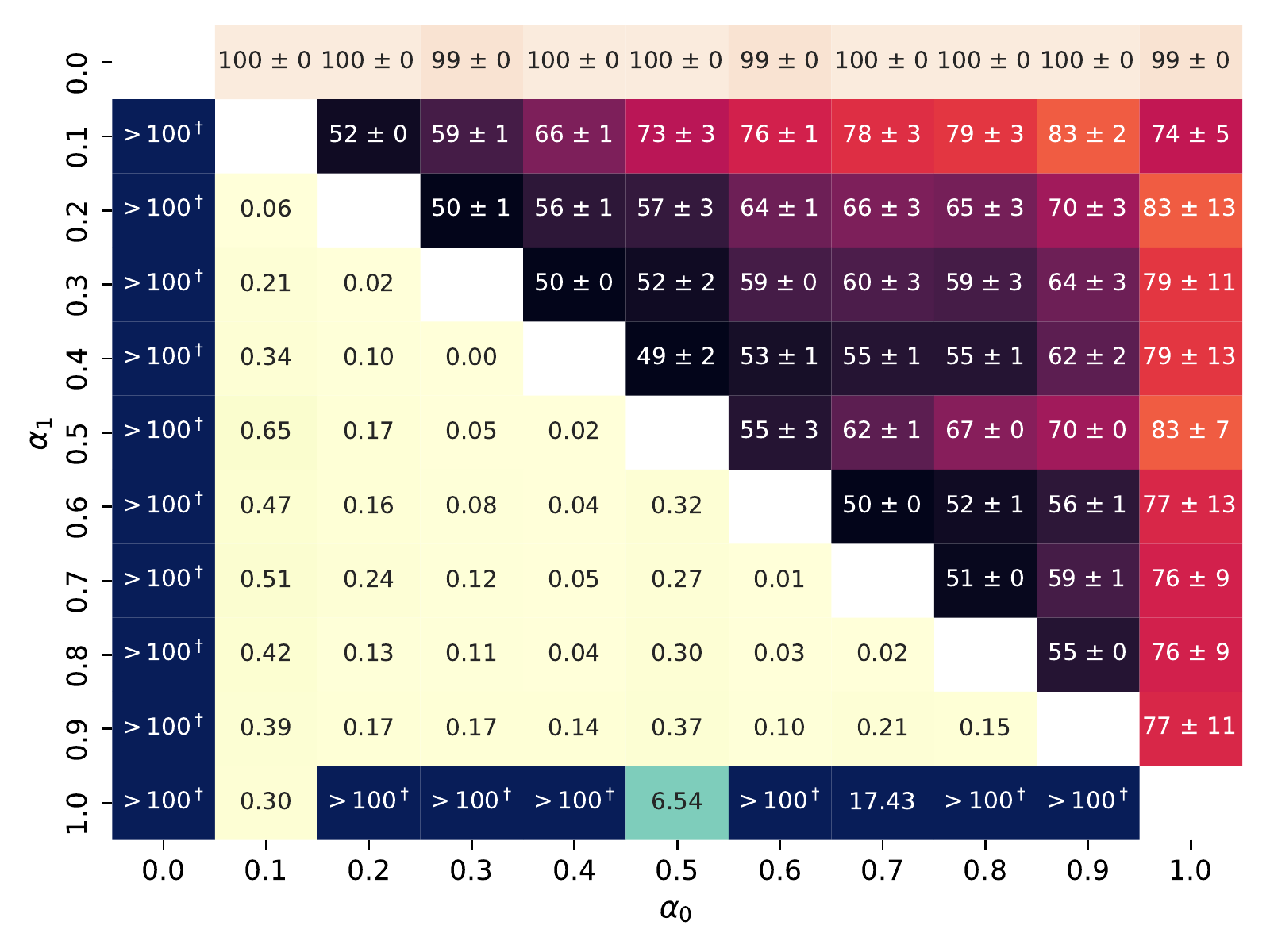}
\caption{Census}\label{fig:heatmap_census}
\end{subfigure}
\hfill
\begin{subfigure}[b]{0.49\textwidth}
    \centering
    \includegraphics[width=1.0\textwidth]{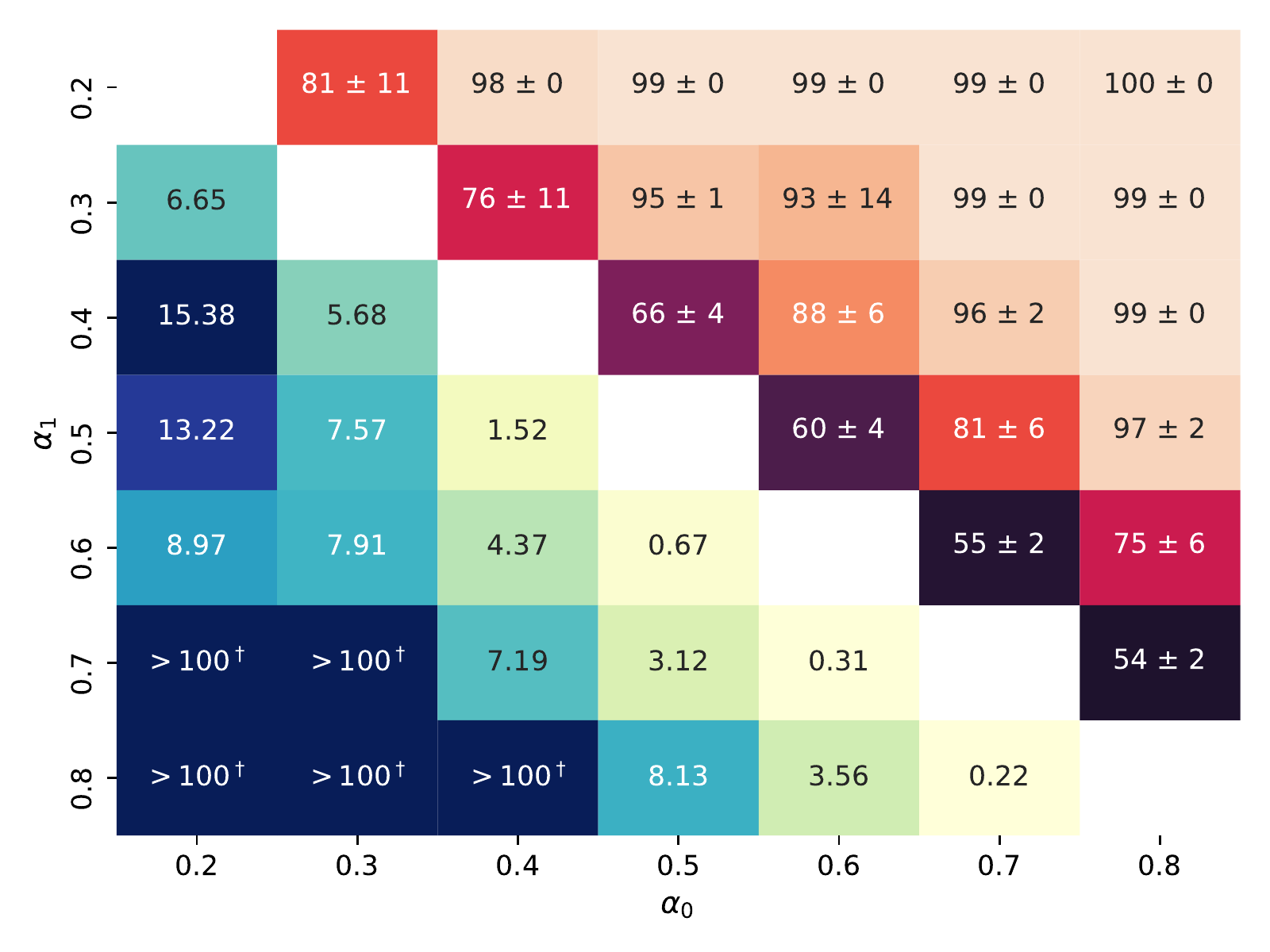}
    \caption{\boneage}
    \label{fig:boneage_heatmap_meta}
\end{subfigure}
\caption{
Effectiveness of meta-classifiers in distinguishing proportions of females on \census\ and \boneage\
.
The bottom-left triangles of the heatmaps show the \neff\ values, and the top-right triangles show the distinguishing accuracies between training distributions \gzero($\mathcal{D}$) with ratio $\alpha_0$ and \gone($\mathcal{D}$) with ratio $\alpha_1$ for females. Distinguishing accuracies seem to follow intuitive patterns, with an increase as the distributions diverge (larger $|\alpha_0 - \alpha_1|$). The \neff\ values allow for comparisons of attack power between different pairs of distributions, but also show that very little leakage is observed for most settings, except for \boneage.}
\label{fig:heatmap_census_boneage}
\end{figure*}

\subsubsection{Varying Proportions}
\label{sec:finegrained}

An adversary may not necessarily be interested in distinguishing between the balanced case ($\alpha=0.5$) and other ratios. For instance, health datasets for specific ailments may have a higher underlying prevalence in females, and the adversary may be interested in differentiating between two particular ratios of females, like $0.3$ and $0.4$. We thus experiment with the case where both distributions are varied: \gzero($\mathcal{D}$) and \gone($\mathcal{D}$) with corresponding ratios $\alpha_0$ and $\alpha_1$ respectively. 
Distribution inference risk seems particularly acute when an adversary can distinguish the proportion of an uncommon property. Observing performance trends as the difference in ratios increases also helps us understand how much of a threat property inference may pose as the similarity of the distributions varies. 

Figure~\ref{fig:heatmap_census_boneage} shows the distinguishing accuracies (and corresponding \neff\ values) between models (in the form of heatmaps) trained on distributions \gzero($\mathcal{D}$) and \gone($\mathcal{D}$) while varying corresponding $\alpha_0$ (horizontal axis) and $\alpha_1$ (vertical axis), for ratios of females on \census and \boneage. For instance, in Figure~\ref{fig:heatmap_census}, $(\alpha_0, \alpha_1)=(0.2,0.9)$ in the upper-right triangle correspond to meta-classifier performance ($70\%$), while $(\alpha_0, \alpha_1)=(0.9,0.2)$ in the lower-left triangle gives the corresponding $\neff = 0.17$.
Entries along a given diagonal have the same value of $|\alpha_0 - \alpha_1|$. As reflected in the heatmap colors,  distinguishing accuracies are roughly the same along diagonals. The variance in performance across runs is relatively high for similar distributions (small $|\alpha_0-\alpha_1|$) and decreases as the distributions diverge. For \celeba\ (sex) and \census\ (sex), we observe that \neff$<1$ in most cases. These small values do not imply the inability of \textit{any} adversary to distinguish between the distributions, only that for the given attacks we observe little information leakage.






\subsection{Direct Regression over $\alpha$} \label{sec:regression_results}

Inspired by Zhou \textit{et al.}~\cite{zhou2021property}, we performed a direct regression experiment in which we trained the meta-classifiers to predict $\alpha$ 
directly. This corresponds to a more realistic attack setting for many scenarios than one in which the adversary is distinguishing between two predefined $\alpha$ values.
For this experiment, we construct a training dataset for the regression meta-classifier with tuples of the form $(M_{\alpha}, \alpha)$, where $M_{\alpha}$ is some model with a training distribution corresponding to the ratio $\alpha$. We train the meta-classifier using $M_{\alpha}$ models for all the ratios $\alpha$ that we experiment with in Section~\ref{sec:finegrained} ($\{0.0, 0.1, \ldots, 1.0\}$ for \census\ and \celeba, and $\{0.2, \ldots, 0.8\}$ for \boneage). The meta-classifier follows the same permutation-invariant architecture as in the binary property experiments (Section~\ref{sec:whiteboxattacks}), just with a mean squared error (MSE) loss for training.

\begin{figure*}[tb]
\centering
\begin{subfigure}[b]{0.48\textwidth}
    \centering
    \includegraphics[width=\textwidth]{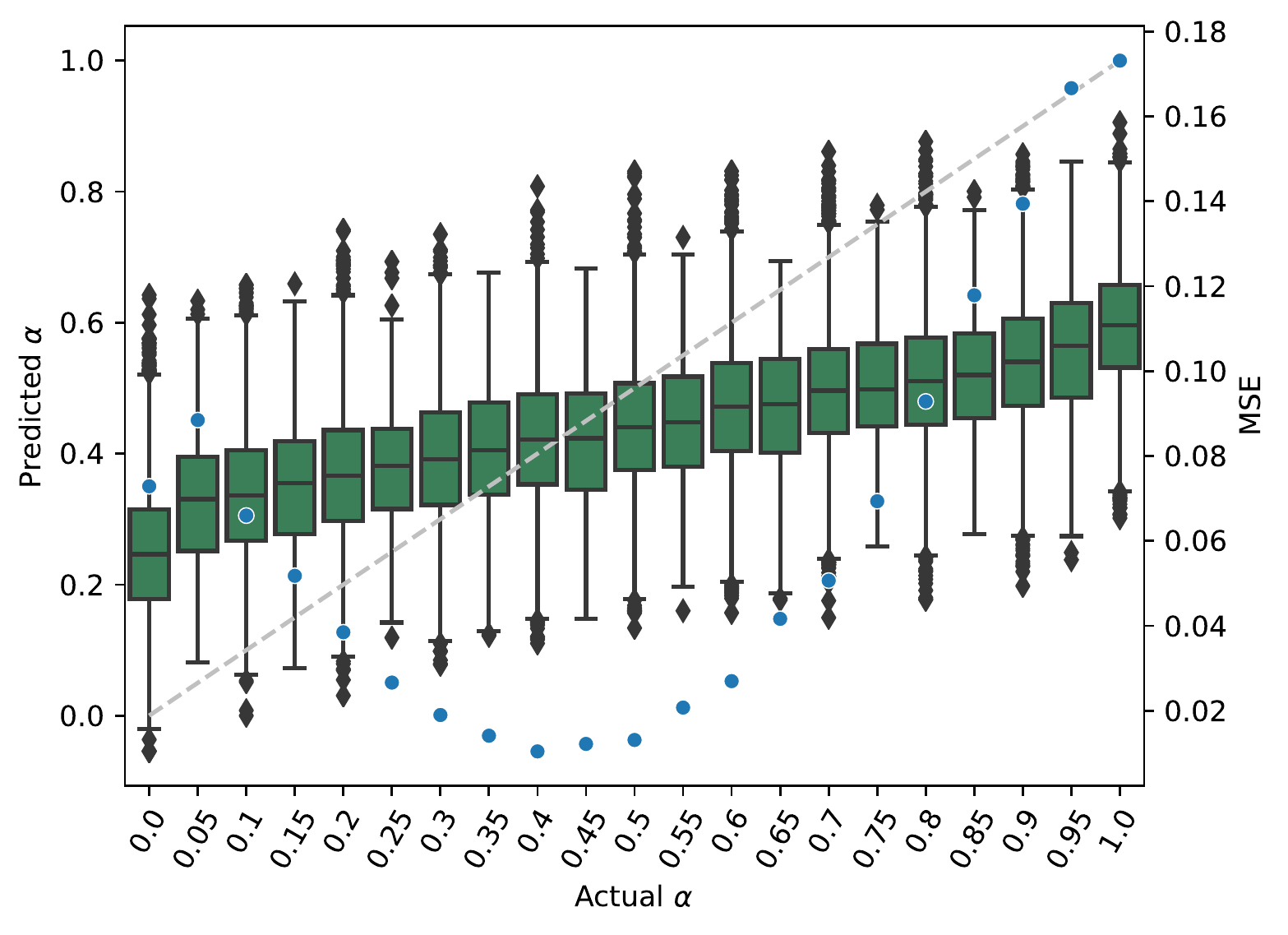}
    \caption{\census\ (Sex)}
    \label{fig:regression_plot_census}
\end{subfigure}
\hfill
\begin{subfigure}[b]{0.48\textwidth}
    \centering
    \includegraphics[width=\textwidth]{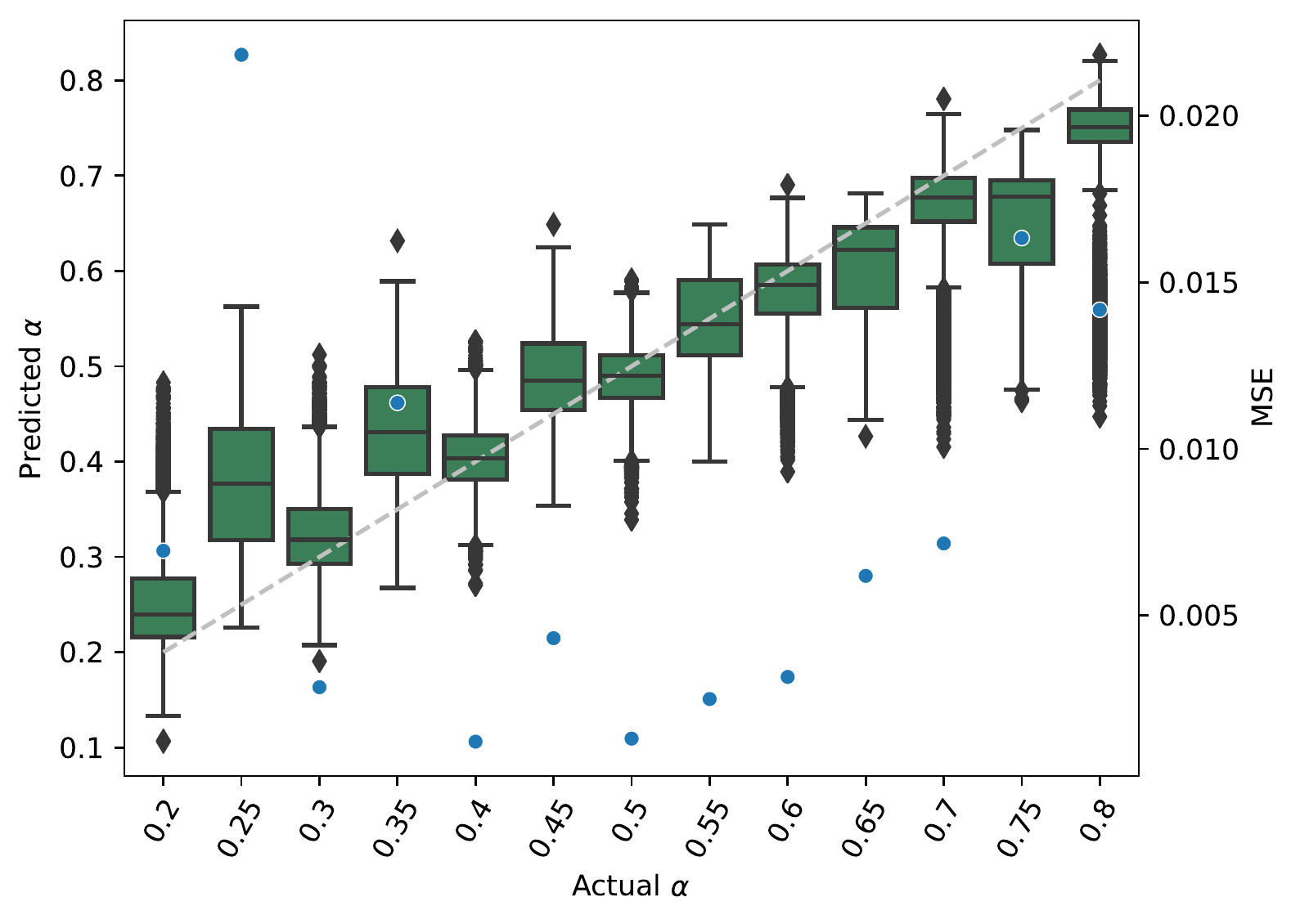}
    \caption{\boneage}
    \label{fig:regression_plot_boneage}
\end{subfigure}
\caption{Predicted $\alpha$ values (left y-axis) for models with training distributions for varying $\alpha$ values (x-axis), for all victim models and regression meta-classifier experiments (green box-plots), along with mean squared error (right y-axis labels, with different scales on the two graphs, and blue dots), for (\protect{\subref{fig:regression_plot_census}}) \census\ and (\protect{\subref{fig:regression_plot_boneage}}) \boneage\ datasets. with The diagonal gray dashed line represents the ideal case, where the regression classifier would perfectly predict $\alpha$. For each ratio of the form $0.05 \cdot x$ (for varying $x$), we train regression meta-classifiers 5 times with different seeds, and test 100 victim models. As indicated by \neff\ values, the \boneage\ dataset observes very good performance, with nearly all predictions lining up with the diagonal, while for \census\ (sex), predicted ratios are usually in $[0.2, 0.6]$.}
\label{fig:regression_plots}
\end{figure*}

\begin{table}[t]
\small
\centering
\begin{tabular}{c c c c c c} 
 \toprule
 \multirow{2}{*}{Dataset} & \multirow{2}{*}{Attribute} & \multicolumn{3}{c}{\neff} & \multirow{2}{*}{MSE} \\
 & & (B) & (BR) & (R) & \\
 \midrule
 \multirow{2}{*}{\census} & Sex & 0.2 & 0.3 & 8.8 & 0.053\\
 & Race & 0.1 & 0.2 & 6.0 & 0.091\\
 \midrule
 \multirow{2}{*}{\celeba} & Sex & $0.3$ & $0.4$ & 10.6 & 0.030\\
 & Age & 0.2 & 0.4 & 5.1 & 0.047\\
 \midrule
 \boneage & Sex & 6.2 & 15 & 269.4 & 0.001\\
 \bottomrule
\end{tabular}
\caption{Median \neff\ values when using the binary meta-classifiers \neff\ (B), regression meta-classifiers \neff\ (R), and regression meta-classifiers for binary predictions \neff\ (BR), along with average Mean Squared Error (MSE) for direct regression over $\alpha$. \neff\ is nearly double for all cases that use regression meta-classifiers for binary predictions, when compared to the binary meta-classifiers.}
\label{table:regression_for_binary}
\end{table}

Figure~\ref{fig:regression_plots} shows the distribution of the predictions of the regression meta-classifiers and Table~\ref{table:regression_for_binary} reports the MSE and \neff\ values. For all these experiments, we train meta-classifiers five times with different seeds, and report aggregate results over all the meta-classifiers and victim models, for each dataset and property.  In the plots in Figure~\ref{fig:regression_plots}, we include results for actual $\alpha$ values at both tenths and the intermediate $0.05$ ratios to confirm that the meta-classifiers are indeed learning to predict $\alpha$ and not just overfitting to the $\alpha$ values observed in training. The meta-classifiers do exhibit a bias toward balanced predictions, showing a smooth curve for the MSE values with a minimum near $\alpha = 0.5$. 

The attacks are quite successful in most cases, achieving \neff\ over $5$ for all of our settings, and surprisingly high leakage for \boneage, achieving \neff\ over $260$. These regression attacks show that adversaries can infer sensitive information about training datasets even in the more realistic settings where the adversary does not have prior knowledge of the distributions to distinguish. However, the attacks are not always highly successful--- the performance observed for our meta-classifiers on \census\ (race) is not much better (Figure~\ref{fig:regression_plot_census}) than that when guessing $\alpha=0.5$ blindly (expected MSE $0.1$).

\begin{figure*}[bt]
\centering
\begin{subfigure}[b]{0.49\textwidth}
\includegraphics[width=0.9\textwidth]{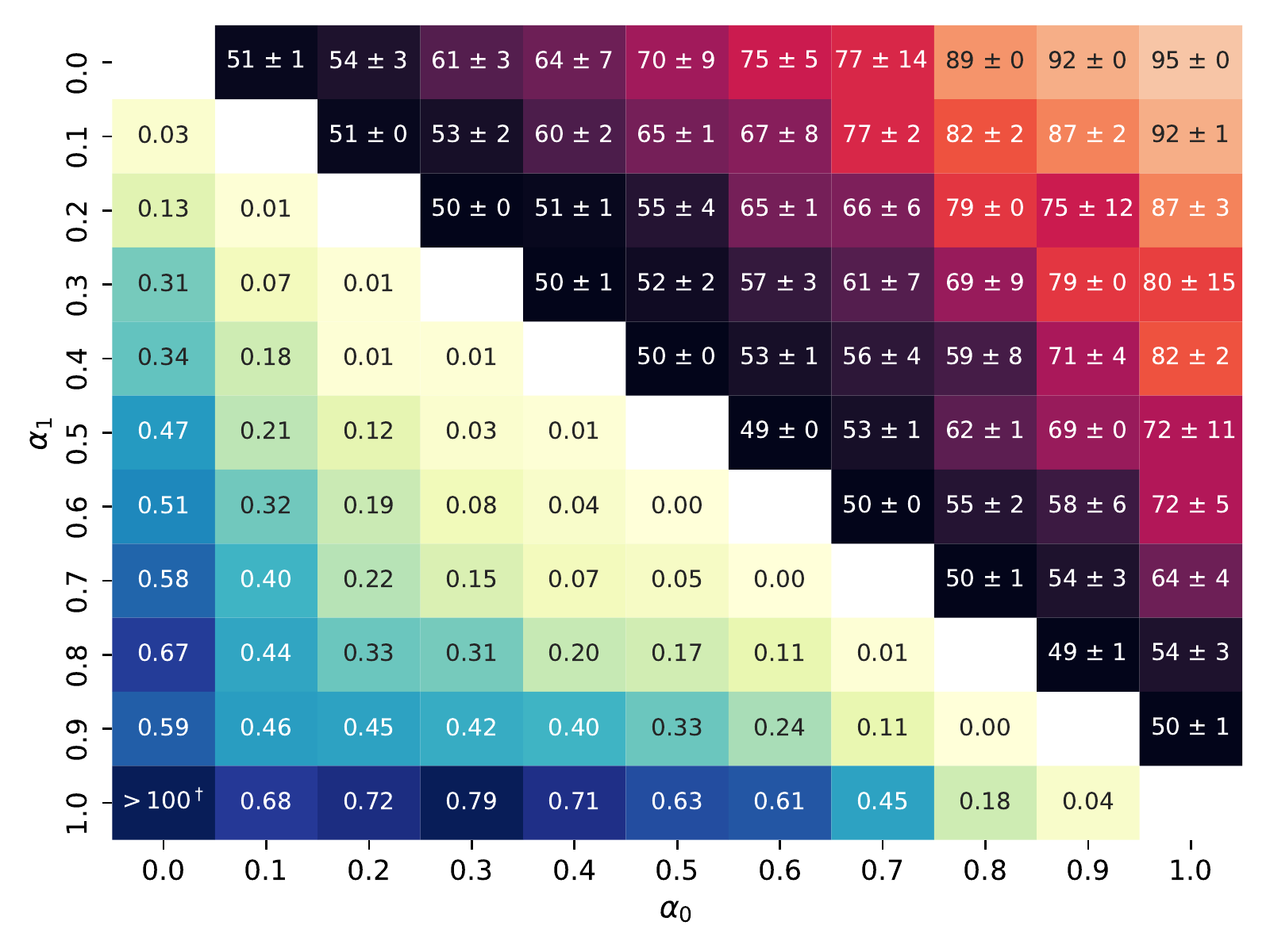}
\caption{Binary Classifiers}\label{fig:heatmap_celeba_binary}
\end{subfigure}
\hfill
\begin{subfigure}[b]{0.49\textwidth}
\includegraphics[width=0.9\textwidth]{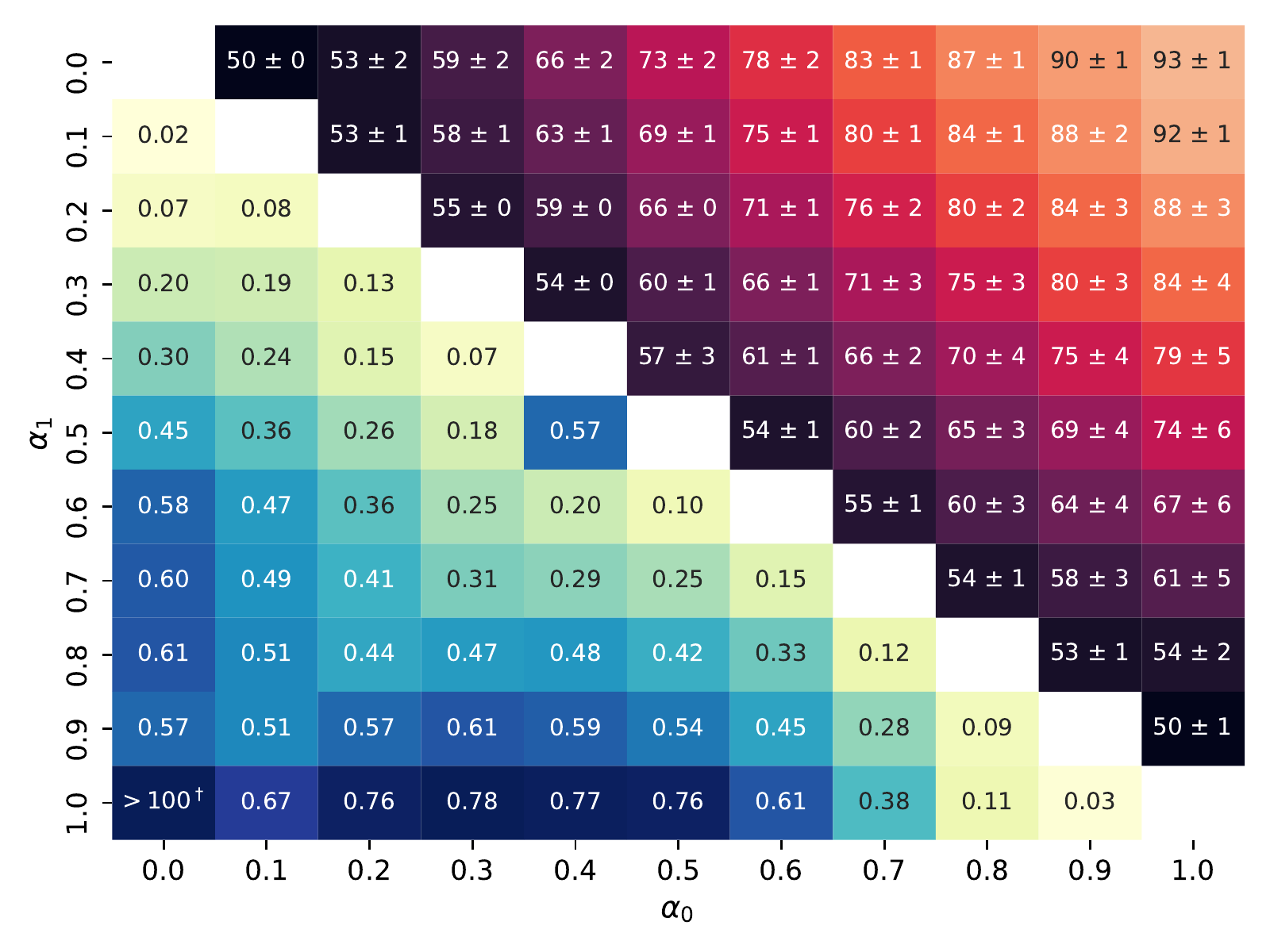}
\caption{Regression Meta-Classifier}\label{fig:heatmap_celeba_regress}
\end{subfigure}
\caption{Distinguishing binary ratio properties using (\subref{fig:heatmap_celeba_binary}) binary classifiers and (\subref{fig:heatmap_celeba_regress}) regression meta-classifiers, for \celeba\ (age). \neff\ values (lower triangle) and  classification accuracies (upper triangle) for distinguishing proportion of old people (ratios $\alpha_0$, $\alpha_1$) in training data for the \celeba\ dataset. \neff\ values are higher than those observed for binary meta-classifiers, especially for cases where $|\alpha_0 - \alpha_1|$ is small.}
\label{fig:heatmap_celeba_compare}
\end{figure*}

Given the high \neff\ values for the regression tests, we tried using the regression meta-classifiers to distinguish between binary properties. To produce a classification between models with training distribution ratios $\alpha_0$ and $\alpha_1$, a regression meta-classifier $M_{\text{regression}}$'s prediction for some model $m$ is converted to a binary outcome by simply checking which of the two considered distribution ratios the predicted ratio is closer to: 
$    \hat{b} = \mathbb{I}\left[M_{\text{regression}}(m) \geq \frac{\alpha_0 + \alpha_1}{2}\right]$. 
Each entry in Table~\ref{table:regression_for_binary} is averaged over 5 trials $\times$ 100 victim models $\times$ 11 ratios ($\{0.0, 0.1, \ldots, 0.9, 1.0\}$ for \census; 7 in the case of \boneage). 
In most cases, the accuracy improves significantly over the binary classifiers, with the \neff\ value nearly doubling for most settings. For instance, the classification accuracy increases by $\sim4\%$ and $\sim15\%$ for \celeba\ (sex) and \boneage\ respectively,
corresponding to an increase of 
\neff\ by $\sim0.14$ for \celeba\ (sex) and $\sim8.51$ for \boneage. This improvement is not surprising since the binary attack uses models only from two distributions, whereas the regression attack has models from a wide range of alpha values and thus can learn more. Further looking at the \neff\ values for each pair of ratios ($\alpha_0$, $\alpha_1$) shows how these improvements are uniform across all pairs of distributions. Figure~\ref{fig:heatmap_celeba_compare} shows the accuracies and \neff\ values for using specific ratio binary classifiers compared with the improved accuracies obtained using the regression meta-classifier.

\subsection{Graph Properties}
\label{sec:graph_results}
Our experiments using both binary and regression classifiers on the graph datasets reveal surprisingly high property leakage. Observed \neff\ values for \graph\ are much higher (100-200 range) than was observed in the experiments on tabular and image datasets (with the exception of \boneage). (For the clustering coefficients on \botnet, we do not have a way to compute \neff, but also see evidence of substantial leakage.)

\begin{figure*}[tb]
\centering
\begin{subfigure}[b]{0.49\textwidth}
    \centering
    \includegraphics[width=0.8\textwidth]{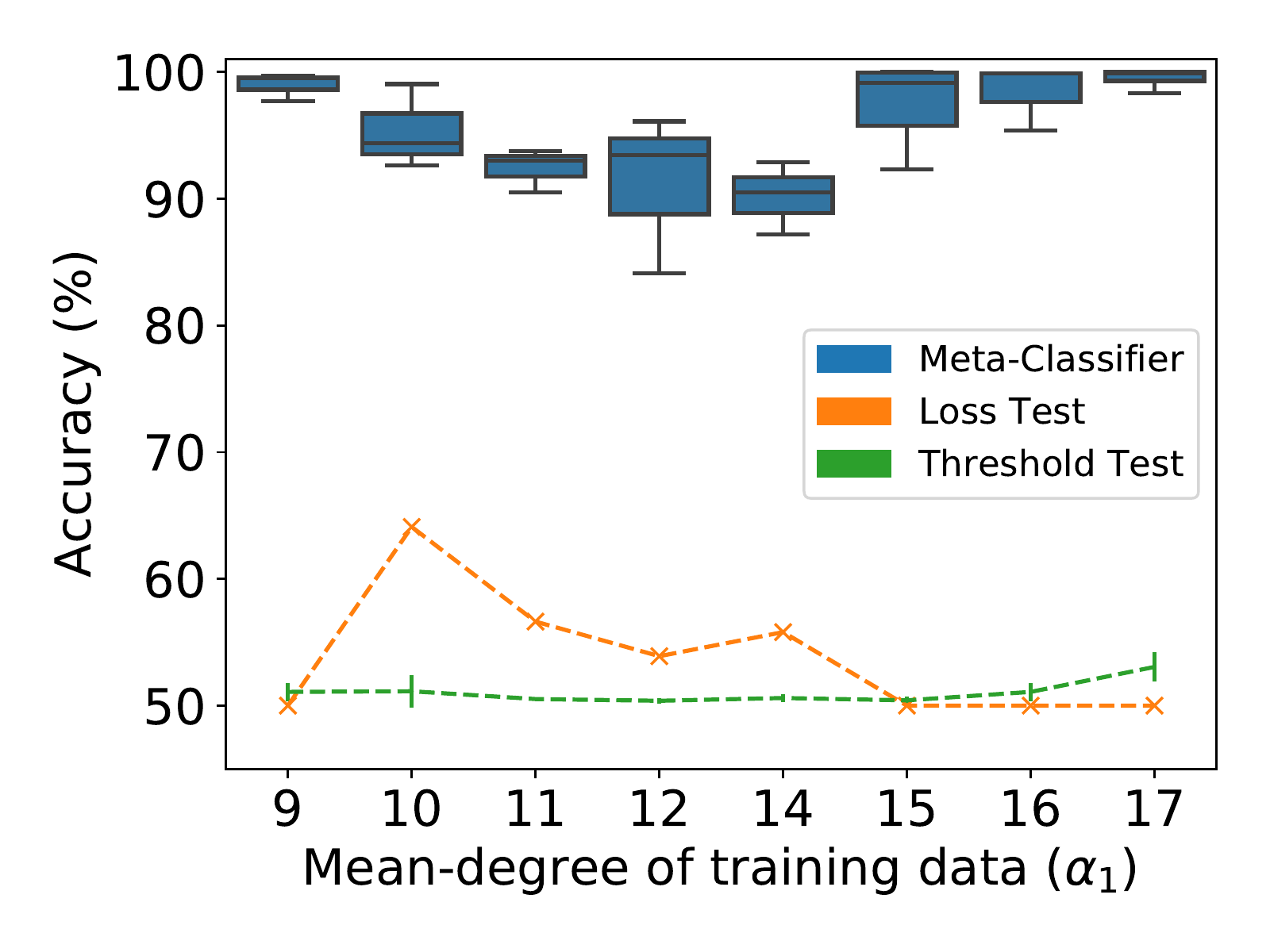}
        \vspace*{-1ex}
    \caption{Distinguishing accuracy ($\alpha_0 = 13$) as mean degree varies.}
\label{fig:arxiv_classify}
\end{subfigure}
\begin{subfigure}[b]{0.49\textwidth}
    \centering
    \includegraphics[width=0.75\textwidth]{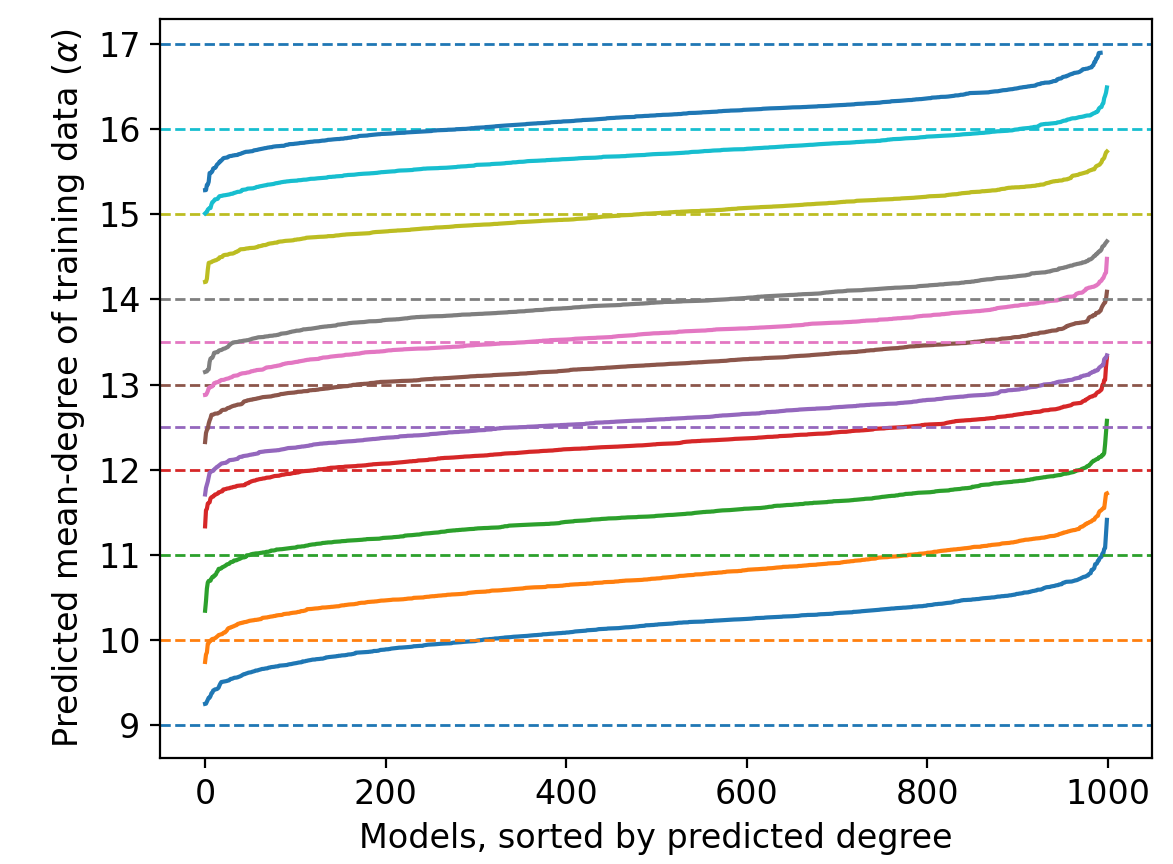}
    \caption{Mean node-degree $\alpha_1$ as predicted by the meta-classifier.}
    \label{fig:arxiv_regress}
\end{subfigure}
\caption{Performance for (\protect\subref{fig:arxiv_classify}) distinguishing between models with different mean node-degrees in training data, and (\protect\subref{fig:arxiv_regress}) directly inferring the mean node-degree of the training data, for the \graph\ dataset. Each color represents the true degree (dashed lines) of the models being tested.
The meta-classifier attack is remarkably successful on this dataset and further accentuates how some attacks can infer underlying properties nearly exactly on some datasets.}
\label{fig:boneage_plots}
\end{figure*}

\begin{figure}[ptb]
\centering
    \includegraphics[width=0.5\columnwidth]{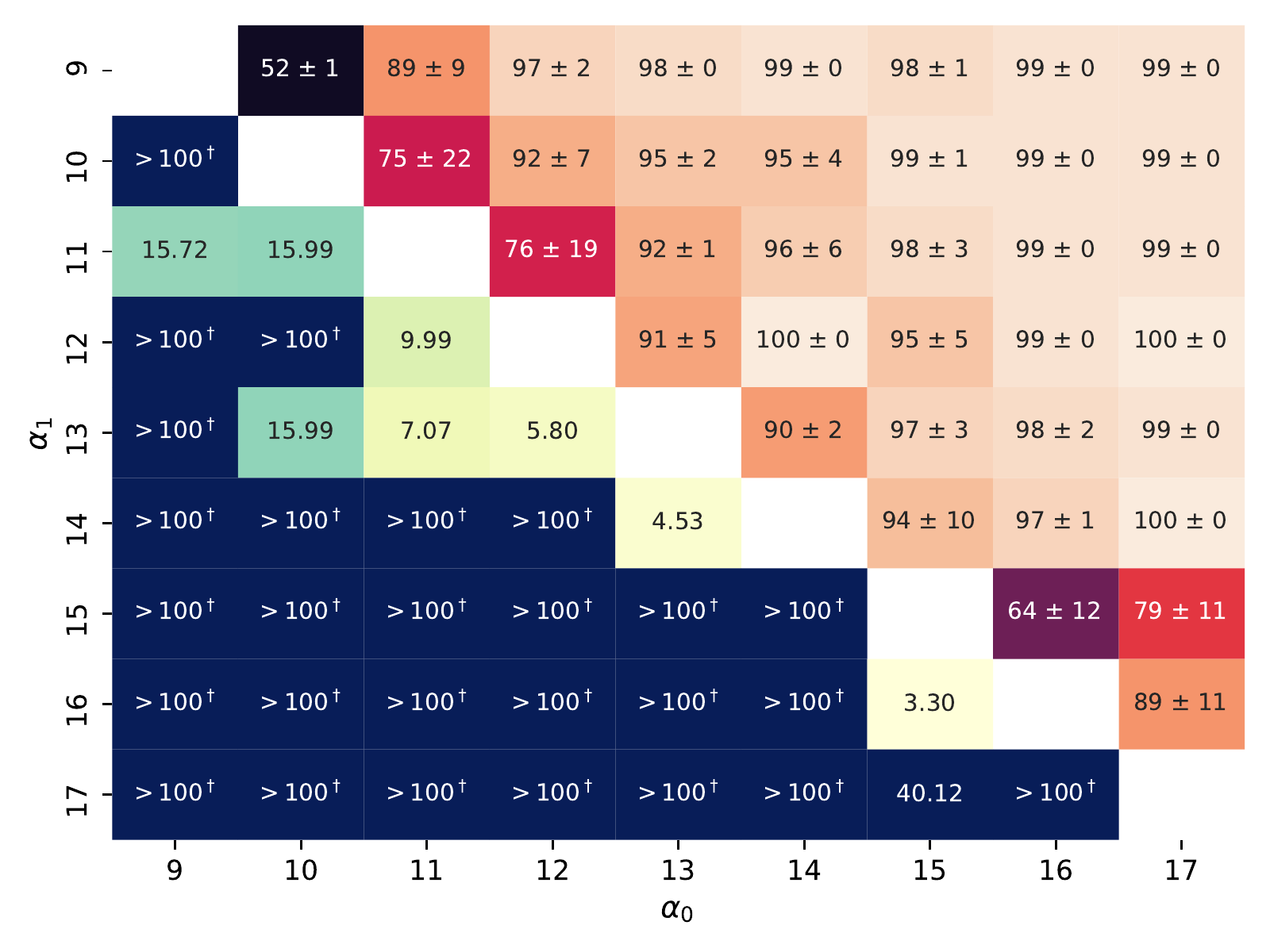}
\caption{
Effectiveness of meta-classifiers on \graph\ dataset. 
\neff\ values (bottom-left triangles) and mean node-degree (degrees $\alpha_0$, $\alpha_1$) in training data.} 
\label{fig:heatmap_ariv_meta}
\end{figure}

\shortsection{\graph}
For the \graph\ dataset, we set \gzero\ such that the graph has a mean node-degree of $\alpha_0=13$, and for \gone, modify the graph to have a mean-degree $\alpha_1$
as an integer in the range [9, 17]. We produce test datasets by pruning either high or low-degree nodes from the original graph to achieve a desired $\alpha_1$. Like the other datasets, meta-classifier performance increases as the distributions diverge, albeit with much smaller drops. Both the \ltest\ and \ttest\ fail on this dataset with \neff\ values below 1, compared to the meta-classifiers (Figure~\ref{fig:arxiv_classify}) which leaks $\neff \approx 40$ in most cases. The attacks leak much more information as the degrees increase than when they decrease--- \neff\ values are nearly double for (12, 14) than (12, 13) as the two mean node-degrees, despite having comparable distinguishing accuracies. Motivated by the success of regression attacks for ratio-based properties (Section~\ref{sec:regression_results}), we also trained a regression variant of the meta-classifier to predict the average degree of the training graph directly. The resulting meta-classifier performs quite well (Figure~\ref{fig:arxiv_regress}), achieving a mean squared error (MSE) loss of $0.393 \pm 0.36$. It generalizes well to unseen distributions, achieving an average MSE loss of 0.076 for $\alpha$ = 12.5 and 13.5. 
A property inference adversary can thus be strong enough to directly predict the average node degree of the training distribution. 
Similar to the case of ratios, we try different combinations of mean node-degree values by setting different mean node-degrees $\alpha_0$ and $\alpha_1$ (Figure~\ref{fig:heatmap_ariv_meta}). As apparent, \neff\ has a wide spread in its values across different distributions---starting from $\approx3$ to approaching infinity, with a median of $\approx31$.

\shortsection{\botnet}
For the \botnet\ dataset, we construct \gzero\ to have graphs with average clustering-coefficient below 0.0061 and \gone\ to have graphs with average clustering-coefficient above 0.0071. We pick these values to minimize the overlap between the two distributions, while maintaining a decent accuracy on the original task.
For the case where both the trainer's and adversary's datasets are sampled from the same pool of data, the adversary has near-perfect distinguishing accuracy, even when training the meta-classifier with ten models (and testing on 1000). However, the adversary cannot achieve the same level of performance in the absence of data overlap. Using the \ltest\ yields an accuracy of 63\%, while the \ttest\ and meta-classifier struggle to perform better than random ($\approx$51\%). This disparity in performance further justifies our experimental design choice to consider non-overlapping data splits---evaluating property inference attacks on models trained from the same dataset pool seems effective, but it cannot distinguish learning some unrelated property from the claimed inference.
Additionally, the gap in meta-classifier performance and the simple \ltest\ suggests there might exist better methods that could infer properties even under the non-overlapping scenario.

\subsection{Summary of Experimental Results}

\begin{figure*}[t]
\centering
\begin{subfigure}[b]{0.48\textwidth}
    \centering
    \includegraphics[width=\textwidth]{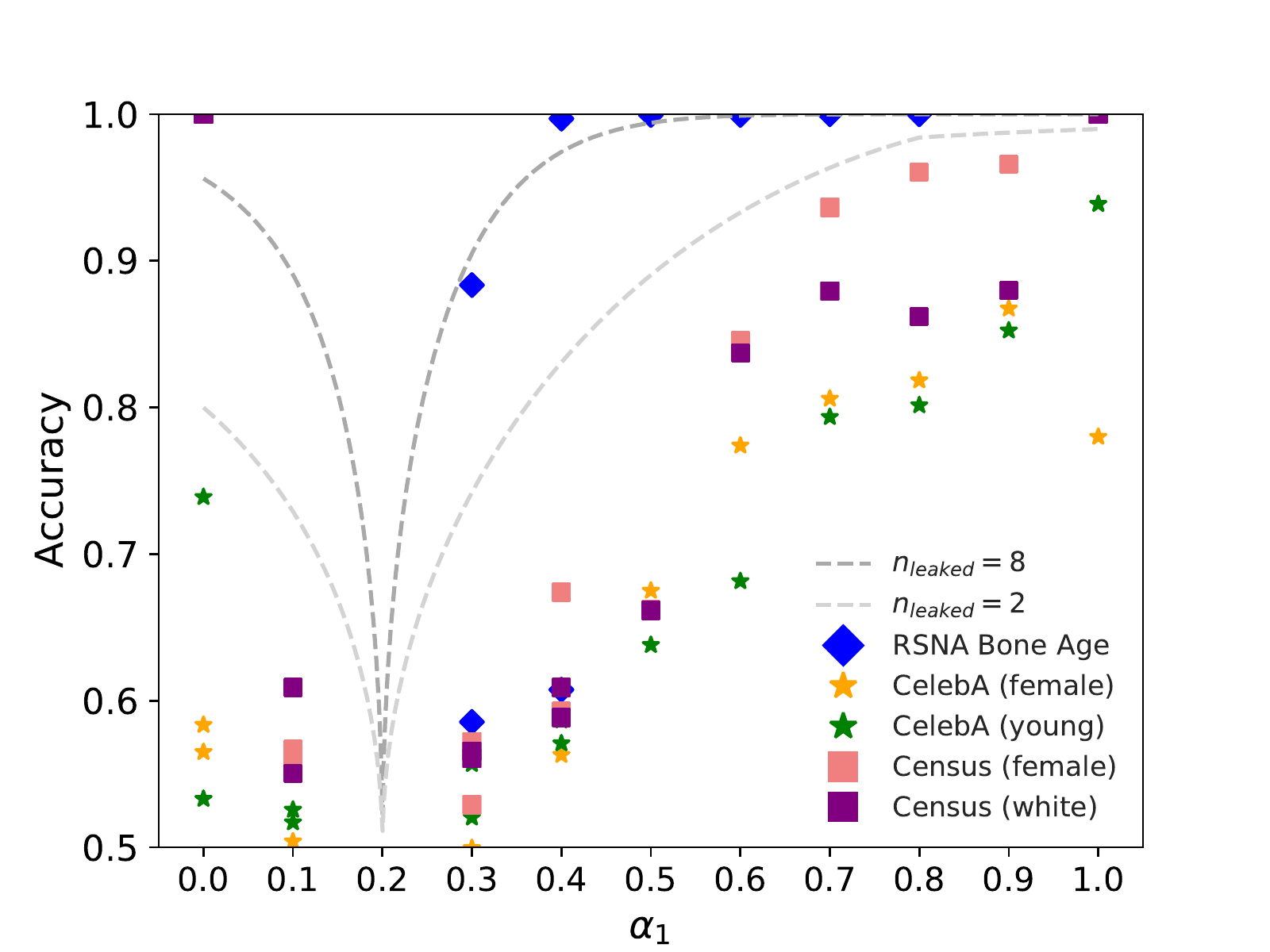}
    \caption{Distinguishing accuracy with varying $\alpha_1$ ($\alpha_0=0.2$).}
    \label{fig:bound_acc}
\end{subfigure}
\hfill
\begin{subfigure}[b]{0.48\textwidth}
    \centering
    \includegraphics[width=\textwidth]{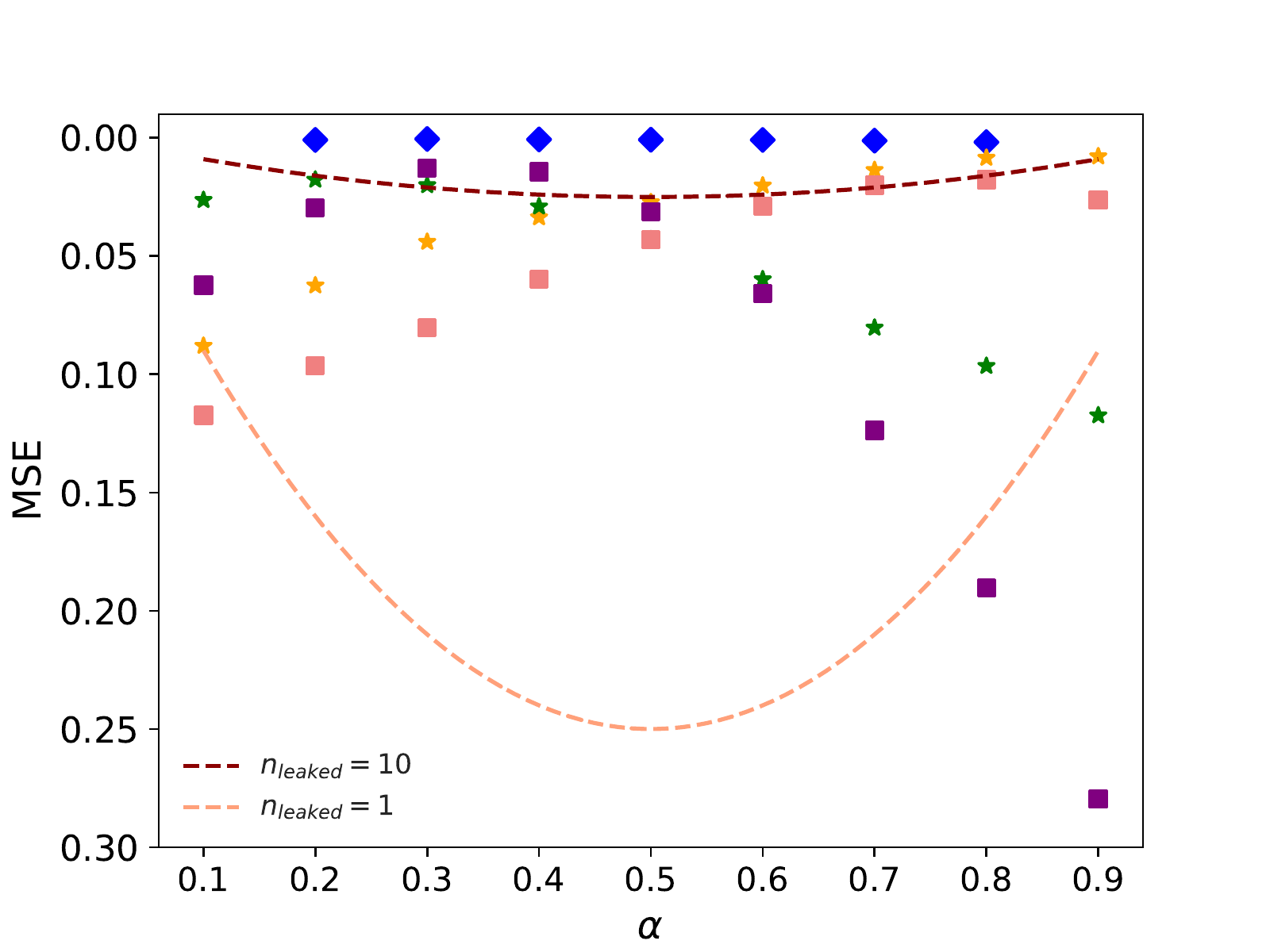}
    \caption{Mean Square Error (MSE) for regression inference.}
    \label{fig:bound_regress}
\end{subfigure}
\caption{Distinguishing accuracy (\protect\subref{fig:bound_acc}) and MSE (\protect\subref{fig:bound_regress}) for inference attacks with varying $\alpha_1$ on the horizontal axis. The plotted results are for the most effective attacks from the experiments described in Section~\ref{sec:experiments}. The curves in (\protect\subref{fig:bound_acc}) show the comparable distinguishing accuracy for $\neff = 2$ (indicating that most of the attacks are comparable to leaking fewer than two samples from the training distribution) and $\neff = 8$, showing that a few of the attacks on the \boneage\ dataset (and extreme attacks on \census\ for the race attribute) do leak a substantial amount of information. Similar trends hold for regression, with \neff\ somewhere between 1 and 10 for most cases (\protect\subref{fig:bound_regress}). The highest leakages we observed are for the graph datasets, not shown in this figure.
}
\label{fig:bound_curves_both}
\end{figure*}

We summarize the distribution leakage observed for our ratio and regression based experiments in Figure~\ref{fig:bound_curves_both}. The results show that the most effective property inference attacks for all datasets (other than \boneage) are less accurate predictors than would be possible with the best possible statistical test from a random sample of size two from the training distribution. Most attacks across varying distributions and datasets correspond to values of $\neff \approx1$ for classification, and $\neff$ between 5 and 10 for regression. The low $\neff$ values for binary classifiers show how little information those attacks are able to leak from the model, but the regression results do indicate that even in these settings non-trivial information is leaking and there are opportunities for better inference attacks. The biggest exception is for \boneage, where we observe \neff\ values above 10 for the binary classifiers (Figure~\ref{fig:boneage_heatmap_meta}) and up to 270 for the regression meta-classifier, and the graph datasets, where they are in the hundreds for \graph\ (Figure~\ref{fig:heatmap_ariv_meta}). 

\shortsection{Dataset Inference Susceptibility} This leakage is much higher for \boneage\ than it is for similar Boolean ratio-like properties for datasets like \census\ and \celeba, suggesting this particular dataset is prone (much more than the others we explore) to high-confidence inference attacks, at least for the sex property. We do not yet have a good understanding of why this dataset leaks so much more distribution information than the other two, but think the causes of variation in inference risk is an important question for future study.
Although we observe an increase in \neff\ values across all our experiments as the distributions diverge from each other, these trends are not symmetric. Hence, not only do more divergent distributions seem to leak more information, that information is more valuable to an adversary since as the distributions diverge less information is needed to distinguish them accurately. This might be suggestive of biased learning algorithms, or perhaps the attacks' capabilities to infer distributional properties effectively. 

\shortsection{Attack Comparison} 
Analyzing \neff\ values for different attacks, which is one of the primary purposes of that notion, shows there is no clear winner out of the three attacks in our experiments. The concept of \neff\ serves as a measure for comparing the power of different attacks across ranges of distributions---seemingly different distinguishing accuracies can correspond to very close \neff\ values (for instance, \celeba), while similar ones can correspond to very different \neff\ values (for instance, \graph). The lack of any clear ranking of the simple black-box and white-box meta-classifier attacks shows why it is necessary to include simple baselines when evaluating property inference risk. In fact, the inexpensive \ttest\ does better than meta-classifiers for about $30\%$ of our experiments across all the datasets. There are many instances where either of the two simple black-box tests perform exceptionally well---even better than meta-classifier. For instance, \ttest\ on \census\ (race) and \ltest\ on \celeba\ (old) outperform the meta-classifiers for nearly all of pairs of ratios. These observations further support the approach of using more direct attacks like \ltest\ and \ttest\ before training expensive meta-classifiers, especially since there is no clear ordering of these attacks across datasets and ratios, and also considering attacks that combine meta-classifiers with direct loss tests to improve overall accuracy.

\shortsection{Peculiar Trends}
We also observe some trends specific to a dataset and property, but can only speculate on their causes. For example, on the \census\ dataset, the adversary has notably high accuracy in differentiating between distributions when one is without any females ($\alpha_0=0$) or males ($\alpha_0=1$) with distinguishing accuracies close to 100\%, regardless of the actual proportion of females in the data.
This suggests that detecting the mere presence or absence of members with a particular attribute is significantly easier than trying to deduce the exact ratio of members with that attribute, and perhaps is unsurprising here for an attribute that does impact the task predictions. Similarly, a difference in ratios of $\geq 0.3$ on \boneage\ (Figure~\ref{fig:boneage_heatmap_meta}) yields at least 90\% accuracy for all cases using meta-classifiers, with \neff\ values $\geq7$, going up to perfect distinguishing accuracy.
Unlike \census, performance on \celeba\ at the extremes (no males or females when inferring sex ratios, and no young or old people when inferring old ratios) is far from perfect. This may be because features like race and gender in \census\ are directly used for model training, and thus their presence or absence would directly impact both predictions and model parameters. Whereas for \celeba, the complicated feature extractor may not embed these latent (and inherently ambiguous) properties explicitly.

\shortsection{Regression for Binary Classification}
Comparing \neff\ values for the binary meta-classifiers and regression-based meta-classifiers tuned for binary classification demonstrates how additional information about the underlying ratios can have a huge impact on leakage. Having models trained on training distributions for a wide range of $\alpha$ can help ensure the meta-classifier actually learns to infer the underlying ratios, compared to the binary classification case where it is most likely to rely on specific signals just to distinguish between two given distributions. Although that is indeed the given task, the ability to capture the association between $\alpha$ and the desired predictions can, and does, help the meta-classifiers improve their performance. Note that training the regression-based meta-classifiers does not require a stronger threat model than is assumed for the binary classifier case. In both cases, the adversary needs access to a training distribution with enough samples to be able to create representative datasets for different distributions. Training the regression meta-classifier requires more computational resources (training models for multiple ratios) than is required to train the binary meta-classifier (training models only for two ratios), but does not otherwise require a stronger adversary.

\section{Conclusions}
\label{sec:conclusion}

A important step to developing understanding of distribution inference risks is a precise and formal definition, and we found the general definition we introduce to be useful for conceptualizing the space of attacks especially in having a precise way to separate intended statistical inference from distribution inference attacks. The definition also leads to a systematic approach to quantifying the leakage from distribution inference attacks. Our empirical results reveal how intuition may not necessarily align with actual observations. Seemingly similar pairs of distributions can have starkly different attack success rates, and simple attacks with limited access can sometimes outperform computationally expensive meta-classifiers. Our experiments also show how direct regression of underlying ratios of training distributions is a real threat, and can be used to improve the performance of binary distinguishing attacks.

Our work raises more questions than it answers---\emph{why do some models leak a lot of information about certain properties of their training distribution but others leak little}, \emph{what are the limits on how precisely training distributions can be distinguished}, \emph{why do models trained on some datasets (like \boneage\ and the graphs) appear to leak so much more information than others}. We are not able to answer these questions yet, although our experiments provide several intriguing observations and suggest possibilities to explore. It is not surprising that so little is understood about distribution inference---the research community has put extensive effort into studying membership inference attacks for several years now, and we are just beginning to be able to understand how and why membership inference risk varies~\cite{kulynych2022disparatevulnerability}. There could also be trade-offs between accuracy, robustness, fairness, interpretability, and vulnerability to distribution inference attacks. For instance, a model that is fair with regards to its predictions across some sensitive attribute may be less vulnerable to distribution inference attacks on that attribute.

Another aspect of these attacks, which is accounted for theoretically but hard to implement in practical attacks, is robustness to different training processes. It is unclear how factors like overfitting, training for robustness, or data augmentation can impact property inference risk. Exploring how these factors increase or decrease susceptibility is part of ongoing work and may pave the way for understanding these attacks better, perhaps even leading to principled and effective defenses.

We expect there is room for improving distribution inference attacks, and hope our results raise awareness that distribution inference attacks that expose sensitive aspects of training data are possible and require further exploration, analysis, and development of mitigations.

\subsection*{Acknowledgements}
\noindent
The authors thank Yifu Lu and Justin Chen for contributing to some of the experiments in this paper, along with Xiao Zhang, Bargav Jayaraman, and Ambar Pal for helpful feedback on the paper including a detailed reading of our proof. This work was partially supported by grants from the National Science Foundation (\#1717950 and \#1915813), and funding from Lockheed Martin Corporation.

\subsection*{Availability}
\noindent
Code
for reproducing our experiments is available at:
\url{https://github.com/iamgroot42/FormEstDistRisks}.

\bibliographystyle{IEEEtran}
\bibliography{references}

\onecolumn
\clearpage
\appendix
\section{Proofs}

\subsection{Proof of Lemma \ref{lmm:distbound}}
\label{sec:appendix_theorem}

Assume the adversary can fully recover a dataset $S$ (of size $n$) from some model $M$ trained on it. Assume $\psi(\cdot)$ is an estimator for testing the hypothesis, i.e. $\psi(S) = b_{\in \{0,1\}}$ means that $S$ comes from $\mathcal{G}_b(\mathcal{D})$. Assuming an equal likelihood of the chosen dataset $S$ being from either distributions, we have:
\begin{align}
    \nonumber \text{Error} &= \frac{1}{2}\big(\text{Pr}_{S\leftarrow \mathcal{G}_0(\mathcal{D})^n} \big[\psi(S) = 1\big] + \text{Pr}_{S\leftarrow \mathcal{G}_1(\mathcal{D})^n} \big[\psi(S) = 0\big]\big) \\ \nonumber
    &= \frac{1}{2}(\text{Type I Error} + \text{Type II Error}).
\end{align}
Combining with the result from~\cite{72730}:
\begin{align}
    \text{Error} &\geq \frac{1}{2} - \frac{1}{2}\delta(\mathcal{G}_0(\mathcal{D})^n, \mathcal{G}_1(\mathcal{D})^n) \\
    \Rightarrow \text{Accuracy} &\leq \frac{1}{2} + \frac{1}{2}\delta(\mathcal{G}_0(\mathcal{D})^n, \mathcal{G}_1(\mathcal{D})^n) \label{eq:acc_bound_proving},
\end{align}
where $\delta()$ is the total variation distance between two probability measures, and $\mathcal{G}_{b\in\{0,1\}}(\mathcal{D})^n$ refers to the distribution of $n$ samples from $\mathcal{G}_{b\in\{0,1\}}(\mathcal{D})$.. Thus, the maximum accuracy while differentiating between datasets sampled from either distribution is bounded by the total variation distance between them.
Let $\rho_b(x)$ be the generative probability density function for some sample $x$ drawn from $\mathcal{G}_b(\mathcal{D})$, for $b \in \{0,1\}$. This density function can then be broken down into a multinomial distribution and priors as:
\begin{align}
    \rho_b(x) = (1-\alpha_b) p_b(\textbf{x}|0) + \alpha_b p_b(\textbf{x}|1),
\end{align}
where $\alpha_b$ is the prior for $p(1)$ corresponding to $\mathcal{G}_b(\mathcal{D})$, and $\rho_b(x|1)$ is the associated conditional generative probability density function. Note that $p_0(\textbf{x}|0)=p_1(\textbf{x}|0)$ and $p_0(\textbf{x}|1)=p_1(\textbf{x}|1)$, since they both come from the underlying distribution $\mathcal{D}$. Without loss of generality, let $\alpha_0 > \alpha_1$ (we omit the case of same ratios, since that is trivially indistinguishable). Then:
\begin{align}
    \alpha_1 \rho_0(x) - \alpha_0 \rho_1(x) & = \alpha_1((1-\alpha_0) p_0(\textbf{x}|0) + \alpha_0 p_0(\textbf{x}|1)) - \alpha_0 ((1-\alpha_1) p_1(\textbf{x}|0) + \alpha_1 p_1(\textbf{x}|1)) \nonumber \\ 
    & = \alpha_1 p_0(\textbf{x}|0) - \alpha_1\alpha_0p_0(\textbf{x}|0) + \alpha_0\alpha_1p_0(\textbf{x}|1) - (\alpha_0p_1(\textbf{x}|0) - \alpha_0\alpha_1p_1(\textbf{x}|0) + \alpha_1\alpha_0p_1(\textbf{x}|1)) \nonumber \\
    & = (\alpha_1 - \alpha_0)p_0(\textbf{x}|0) \leq 0 \nonumber \\
    \Rightarrow  \frac{\rho_0(x)}{\rho_1(x)} & \leq \frac{\alpha_0}{\alpha_1} 
\end{align}
Using this inequality, the relative entropy (KL divergence) from \gone($\mathcal{D}$) to \gzero($\mathcal{D}$) can be written as:
\begin{align}
    D_{KL}(\mathcal{G}_0(\mathcal{D})\;\|\;\mathcal{G}_1(\mathcal{D})) = \int\rho_0(x)\log\left(\frac{\rho_0(x)}{\rho_1(x)}\right)dx & \leq \int\rho_0(x)\log\left(\frac{\alpha_0}{\alpha_1}\right)dx \\
     & = \log\left(\frac{\alpha_0}{\alpha_1}\right)\int\rho_0(x)dx = \log\left(\frac{\alpha_0}{\alpha_1}\right)
\end{align}
Since the function $f$ is binary, a prior of $\alpha_b$ for $p(f(x)=1)$ implies a prior of ($1-\alpha_b$) for $p(f(x)=0)$. Utilizing this symmetry, we can similarly upper-bound $D_{KL}(\mathcal{G}_1(\mathcal{D})\;\|\;\mathcal{G}_0(\mathcal{D}))$ with $\log\left(\frac{1-\alpha_1}{1-\alpha_0}\right)$. Removing the $\alpha_0 \geq \alpha_1$ assumption and replacing with the max/min of these two appropriately, we get:
\begin{align}
    D_{KL}(\mathcal{G}_0(\mathcal{D})\;\|\;\mathcal{G}_1(\mathcal{D})) & \leq \log\left(\frac{\max(\alpha_0, \alpha_1)}{\min(\alpha_0, \alpha_1)}\right)  \\ \nonumber
    D_{KL}(\mathcal{G}_1(\mathcal{D})\;\|\;\mathcal{G}_0(\mathcal{D})) & \leq \log\left(\frac{1-\min(\alpha_0, \alpha_1)}{1-\max(\alpha_0, \alpha_1)}\right)
\end{align}

From~\cite{kl_multi}, we know that:
\begin{align}
    D_{KL}(\mathcal{G}_0(D)^n\;\|\;\mathcal{G}_1(D)^n) = nD_{KL}(\mathcal{G}_0(D)\;\|\;\mathcal{G}_1(D))
\end{align}

Thus, when using a dataset $S$ of size $|S|=n$, the equivalent KL-divergence can be bounded by:
\begin{align}
    D_{KL}(\mathcal{G}_0(\mathcal{D})^n\|\mathcal{G}_1(\mathcal{D})^n) & \leq n\log\left(\frac{\max(\alpha_0, \alpha_1)}{\min(\alpha_0, \alpha_1)}\right) & \\ \nonumber
    D_{KL}(\mathcal{G}_1(\mathcal{D})^n\|\mathcal{G}_0(\mathcal{D})^n) & \leq n\log\left(\frac{1-\min(\alpha_0, \alpha_1)}{1-\max(\alpha_0, \alpha_1)}\right)
\end{align}

Using the relation between total variation distance and KL-divergence~\cite{tsybakov2009introduction}, we know:
\begin{align}
    \delta(\mathcal{G}_0(\mathcal{D})^n, \mathcal{G}_1(\mathcal{D})^n) & \leq \sqrt{1 - e^{-D_{KL}(\mathcal{G}_0(\mathcal{D})^n\;\|\;  \mathcal{G}_1(\mathcal{D})^n)}} \label{eq:tv_with_kl}\\
    & = \sqrt{1 - e^{-nlog\left(\frac{\max(\alpha_0, \alpha_1)}{\min(\alpha_0, \alpha_1)}\right)}} = \sqrt{1 - \left(\frac{\min(\alpha_0, \alpha_1)}{\max(\alpha_0, \alpha_1)}\right)^n}
\end{align}
Similarly, using $D_{KL}(\mathcal{G}_1(\mathcal{D})\;\|\;  \mathcal{G}_0(\mathcal{D}))$ in the inequality above we get:
\begin{align}
    \delta(\mathcal{G}_0(\mathcal{D})^n, \mathcal{G}_1(\mathcal{D})^n) & \leq \sqrt{1 - e^{-D_{KL}(\mathcal{G}_1(\mathcal{D})^n\;\|\;  \mathcal{G}_0(\mathcal{D})^n)}} \\
    & = \sqrt{1 - e^{-nlog\left(\frac{1 - \min(\alpha_0, \alpha_1)}{1 - \max(\alpha_0, \alpha_1)}\right)}} = \sqrt{1 - \left(\frac{1 - \max(\alpha_0, \alpha_1)}{1 - \min(\alpha_0, \alpha_1)}\right)^n}
\end{align}
Since the function $f()$ is boolean, an adversary can choose to focus on a property value of $0$ or $1$, and infer the ratio of one using the other. Thus, any two ratios $(\alpha_0, \alpha_1)$ can be alternatively seen as $(1 -\alpha_0, 1-\alpha_1)$. Combining the two inequalities above and plugging them back in (\ref{eq:acc_bound_proving}), we get:
\begin{align}
    \text{Accuracy} \leq \frac{1}{2} + \frac{\min\left\{\sqrt{1 - \left(\frac{\min(\alpha_0, \alpha_1)}{\max(\alpha_0, \alpha_1)}\right)^n}, \sqrt{1 - \left(\frac{1 - \max(\alpha_0, \alpha_1)}{1 - \min(\alpha_0, \alpha_1)}\right)^n}\right\}}{2}
\end{align}
Note that the proof of this bound hinges on both the distributions originating from the same underlying distribution $\mathcal{D}$, which is why we use \gzero($\mathcal{D}$), \gone($\mathcal{D}$) instead of some arbitrarily defined distributions $\mathcal{D}_0$, $\mathcal{D}_1$.

\subsection{Proof of Theorem~\ref{thm:n_eff}} \label{app:theorem_followup}

Consider Lemma~\ref{lmm:distbound}: let $\omega$ be the observed distinguishing accuracy for some attack. Let \neff\ be the effective value of $n$ corresponding to the given attack, \textit{i.e}. Equating it with the best distinguishing accuracy for this value of $n$, we can compute \neff.
If $\frac{\min(\alpha_0, \alpha_1)}{\max(\alpha_0, \alpha_1)} \geq \frac{1 - \max(\alpha_0, \alpha_1)}{1 - \min(\alpha_0, \alpha_1)}$:
\begin{align}
    2\omega - 1 & = \sqrt{1 - \left(\frac{\min(\alpha_0, \alpha_1)}{\max(\alpha_0, \alpha_1)}\right)^{n_{leaked}}} \\
    \log(1 - (2\omega -1)^2) & = n_{leaked}\Biggl(\log\left(\frac{\min(\alpha_0, \alpha_1)}{\max(\alpha_0, \alpha_1)}\right)\Biggr) \\
    n_{leaked} & = \frac{\log(4\omega(1-\omega))}{\log\left(\frac{\min(\alpha_0, \alpha_1)}{\max(\alpha_0, \alpha_1)}\right)}
\end{align}
Similarly, for the case of $\frac{\min(\alpha_0, \alpha_1)}{\max(\alpha_0, \alpha_1)} < \frac{1 - \max(\alpha_0, \alpha_1)}{1 - \min(\alpha_0, \alpha_1)}$, we get:
\begin{align}
    n_{leaked} & = \frac{\log(4\omega(1-\omega))}{\log\left(\frac{1 - \max(\alpha_0, \alpha_1)}{1 - \min(\alpha_0, \alpha_1)}\right)}
\end{align}
Combining these two cases, we get:
\begin{align}
    n_{leaked} & = \frac{\log(4\omega(1-\omega))}{\log(\max\left(\frac{min(\alpha_0, \alpha_1)}{\max(\alpha_0, \alpha_1)}, \frac{1 - \max(\alpha_0, \alpha_1)}{1 - \min(\alpha_0, \alpha_1)}\right))}
\end{align}

\subsection{Proof of Theorem \ref{thm:regress_neff}} \label{app:regression}

Assume the adversary can fully recover a dataset $S$ (of size $N$) from some model $M$ trained on it. Let $n_1$ be the number of entries in $S$ that are 1, and $n_0$ 0 such that $n_0 + n_1 = N$. Then, the conditional probability density function of the underlying distribution $\mathcal{D}$ having $\prob{1}=z$ (assume all $z$ are equally likely), given the observed dataset $S$, can be written using the continuous Bayes' rule as:
\begin{align}
    \condprob{z}{S} = \frac{\condprob{S}{z} \prob{z}}{\int_0^1\condprob{S}{x}\prob{x}dx} = \frac{{N \choose n_1}\ z^{n_1}(1-z)^{n_0}}{\int_0^1{N \choose n_1}\ x^{n_1}(1-x)^{n_0}dx} = z^{n_1}(1-z)^{n_0}\frac{\Gamma(n_0+n_1+2)}{\Gamma(n_0+1)\Gamma(n_1+1)}
\end{align}
The above conditional probability is maximized when $z = \frac{n_1}{N}$, i.e. the guessed ratio is the ratio observed in the given sample $S$. Then, we can compute the expected square error over all possible datasets of size $N$, given that the distribution they were sampled from has a proportion value $\alpha$:
\begin{align}
    \expect{(z-\alpha)^2} & = \expect{z^2} + \alpha^2 - 2\alpha\expect{z} \label{eq:regress_init}
\end{align}
We can then compute $\expect{z}$ as:
\begin{align}
    \sum_{n_1=0}^N\frac{n_1}{N}{N \choose n_1} (\alpha)^{n_1}(1-\alpha)^{N-n_1} = \frac{1}{N}\sum_{n_1=0}^Nn_1{N \choose n_1} (\alpha)^{n_1}(1-\alpha)^{N-n_1} = \alpha \label{eq:regress_m}
\end{align}
Similarly, $\expect{z^2}$ can be computed as:
\begin{align}
    \sum_{n_1=0}^N(\frac{n_1}{N})^2{N \choose n_1} (\alpha)^{n_1}(1-\alpha)^{N-n_1} = \frac{1}{N^2}\sum_{n_1=0}^Nn_1^2{N \choose n_1} (\alpha)^{n_1}(1-\alpha)^{N-n_1} = \alpha^2 + \frac{\alpha(1-\alpha)}{N} \label{eq:regress_v}
\end{align}
Plugging (\ref{eq:regress_m}) and (\ref{eq:regress_v}) in (\ref{eq:regress_init}), we get:
\begin{align}
    \expect{(z-\alpha)^2} = \frac{\alpha(1-\alpha)}{N}
\end{align}
Thus, for an observed square error $\omega$ for some attack, \neff\ can be computed as:
\begin{align}
    n_{leaked} = \frac{\alpha(1-\alpha)}{\omega}
\end{align}

\subsection{Proof of Lemma \ref{lmm:distbound_degree}}
\label{sec:appendix_degree_theorem}


We assume that both degree distributions follow Zipf's law, such that the PDF for either of \gzero\ or \gone can be written as
\begin{align}
    \rho_b(x) & = \frac{x^{-s_b}}{H_{N_b, s_b}}
\end{align}
where $H_{N,s}$ is the $N^{th}$ generalized harmonic number of order $s$, $N_b$ corresponds to the maximum degree (with nonzero probability) $\mathcal{G}_B(\mathcal{D})$, and $s_b$ determines the spread of the distribution.
Since the inequality between the total variation distance and accuracy is independent of the underlying distributions, (\ref{eq:acc_bound_proving}) applies in this case too. 

Without loss of generality, let $N_1 \geq N_0$. In that case, the relative entropy from \gzero\ to \gone\ would be undefined, since support(\gone) $\subseteq$ support(\gzero) would be false. Computing the relative entropy from \gone\ to \gzero, we get:
\begin{align}
    D_{KL}(\mathcal{G}_0(\mathcal{D})\;\|\;\mathcal{G}_1(\mathcal{D})) & =
    \sum_{n=1}^{N_1}\rho_0(x)\log\left(\frac{\rho_0(x)}{\rho_1(x)}\right)
\end{align}
Since $\rho_0(x)$ only applies until $N_0$, it evaluates to 0 for $n > N_0$. Substituting:
\begin{align}
    & \sum_{n=1}^{N_0}\rho_0(x)\log\left(\frac{\rho_0(x)}{\rho_1(x)}\right) + \sum_{n=N_0+1}^{N_1}0\cdot\log\left(\frac{0}{\rho_1(x)}\right) = \frac{1}{H_{N_0, s_0}}\left(\sum_{n=1}^{N_0}x^{-s_0}\left(\log\left(\frac{H_{N_1, s_1}}{H_{N_0, s_0}}\right) + (s_1-s_0)\log(x)\right)\right) \\
    & = \frac{1}{H_{N_0, s_0}}\left(H_{N_0, s_0}\log\left(\frac{H_{N_1, s_1}}{H_{N_0, s_0}}\right) + (s_1 - s_0)\sum_{n=1}^{N_0}x^{-s_0}\log(x)\right)  = \log\left(\frac{H_{N_1, s_1}}{H_{N_0, s_0}}\right) + \frac{s_1-s_0}{H_{N_0, s_0}}\sum_{n=1}^{N_0}x^{-s_0}\log(x)
\end{align}
If $s_1 > s_0$:
\begin{align}
     D_{KL}(\mathcal{G}_0(\mathcal{D})\;\|\;\mathcal{G}_1(\mathcal{D})) & \leq \log\left(\frac{H_{N_1, s_1}}{H_{N_0, s_0}}\right) + \frac{s_1-s_0}{H_{N_0, s_0}}\sum_{n=1}^{N_0}x^{-s_0}\log(N_0) \\
     & = \log\left(\frac{H_{N_1, s_1}}{H_{N_0, s_0}}\right) + (s_1-s_0)\log(N_0)
\end{align}
If $s_1 \leq s_0$:
\begin{align}
     D_{KL}(\mathcal{G}_0(\mathcal{D})\;\|\;\mathcal{G}_1(\mathcal{D})) & \leq \log\left(\frac{H_{N_1, s_1}}{H_{N_0, s_0}}\right)
\end{align}
Computing the total variation distance for $n$ samples according to (\ref{eq:tv_with_kl}) for both cases, we get:
\[
\delta(\mathcal{G}_0(\mathcal{D})^n, \mathcal{G}_1(\mathcal{D})^n) \leq 
\begin{cases}
    \sqrt{1-\left(\frac{H_{N_0, s_0}}{H_{N_1, s_1}}N_0^{s_0-s_1}\right)^n},& \text{if } s_1 > s_0\\
    \sqrt{1 - \left(\frac{H_{N_0, s_0}}{H_{N_1, s_1}}\right)^n}              & \text{otherwise}
\end{cases}
\]
Plugging this in (\ref{eq:acc_bound_proving}) to get an upper bound on the distinguishing accuracy.
\begin{align}
    \text{Accuracy} \leq 
    \frac{1}{2} + \frac{\sqrt{1-\left(\frac{H_{N_0, s_0}}{H_{N_1, s_1}}N_0^{(s_0-s_1)\mathbb{I}[s_1 > s_0]}\right)^n}}{2}
\end{align}

\subsection{Proof of Theorem~\ref{thm:n_eff_degree}} \label{app:theorem_degree_followup}

Consider Lemma~\ref{lmm:distbound_degree}: let $\omega$ be the observed distinguishing accuracy for some attack. Let \neff\ be the effective value of $n$ corresponding to the given attack, \textit{i.e}. Equating it with the best distinguishing accuracy for this value of $n$, we get:
\begin{align}
    \omega & = \frac{1}{2} + \frac{\sqrt{1-\left(\frac{H_{N_0, s_0}}{H_{N_1, s_1}}N_0^{(s_0-s_1)\mathbb{I}[s_1 > s_0]}\right)^{n_{leaked}}}}{2} \\
    \log(1 - (2\omega - 1)^2) & = n_{leaked}\cdot\Biggl(\log\left(\frac{H_{N_0,s_0}}{ H_{N_1, s_1}}\right) + (s_0 - s_1)\mathbb{I}[s_1 > s_0]\log(N_0)\Biggr) \\
    n_{leaked} & = \frac{\log(4\omega(1-\omega))}{\log\left(\frac{H_{N_0,s_0}}{ H_{N_1, s_1}}\right) + (s_0 - s_1)\mathbb{I}[s_1 > s_0]\log(N_0)}
\end{align}

\subsection{Leakage by Layers}
\label{sec:layer_imp}

Observing differences in performance when focusing on different network layers of the same model for \celeba\ (Section~\ref{sec:distinguishing_ratios}, Figure~\ref{fig:celeba_meta_boxplots}) raises an interesting question: \emph{how does information leaked vary across model layers?} Understanding and identifying which layers leak the most information can help better understand the distribution inference risks and how to mitigate them, as well as how to make attacks more efficient. Here, we propose a simple test to help the adversary rank layers for value in distinguishing between the given distributions, 
and show how some layers (the first layer, in most cases) seem to capture properties of the training distribution better than others in a given model.
Meta-classifier attacks are expensive and deciphering what they learn is challenging---identifying critical parameters can both improve understanding and lower resource requirements. If we can use just a fraction of the model's parameters, we may be able to achieve comparable inference performance with fewer shadow models and much lower meta-classifier training costs.

\begin{table*}[h!]
\centering
\begin{tabular}{c c c c c c c c}
\toprule
\multirow{2}{*}{\textbf{Dataset}}  & \multicolumn{7}{c}{\textbf{Layer}} \\
  & \textbf{1} & \textbf{2} & \textbf{3} & \textbf{4} & \textbf{5} & \textbf{6} & \textbf{7}\\
  \midrule
 \celeba\ (female) & \textbf{89.2} & 76.7 & 75.8 & 74.2 & 70.0 & 66.7 & 63.3 \\
 \celeba\ (old) & 64.2 & 60 & 60 & 68.3 & \textbf{73.3} & 70 & 66.7\\
 \midrule
 \census\ (female) & \textbf{62.0} & 58.0 & 56.4 & - & - & - & -\\
 \census\ (white) & \textbf{81.7} & 75.0 & 63.3 & - & - & - & -\\
\midrule
\boneage  & \textbf{65.0} & 64.0 & - & - & - & - & -\\ 
\midrule
\graph  & 93.3 & 95.0 & \textbf{98.3} & - & - & - & -\\ 
\botnet & 69.4 & 60.0 & 64.0 & 64.8 & 57.2 & 50.2 & -\\ 
 \bottomrule
\end{tabular}
\caption{Maximum accuracy using layer-identification method. Since the last layer in all of these models is used for classification with a Softmax/Sigmoid activation, the process in Equation~\ref{eq:identify_useful} cannot be applied to the last layer.}
\label{tab:identify_layers}
\end{table*}
\begin{figure*}
\centering
\begin{subfigure}[b]{0.32\textwidth}
    \centering
    \includegraphics[width=\textwidth]{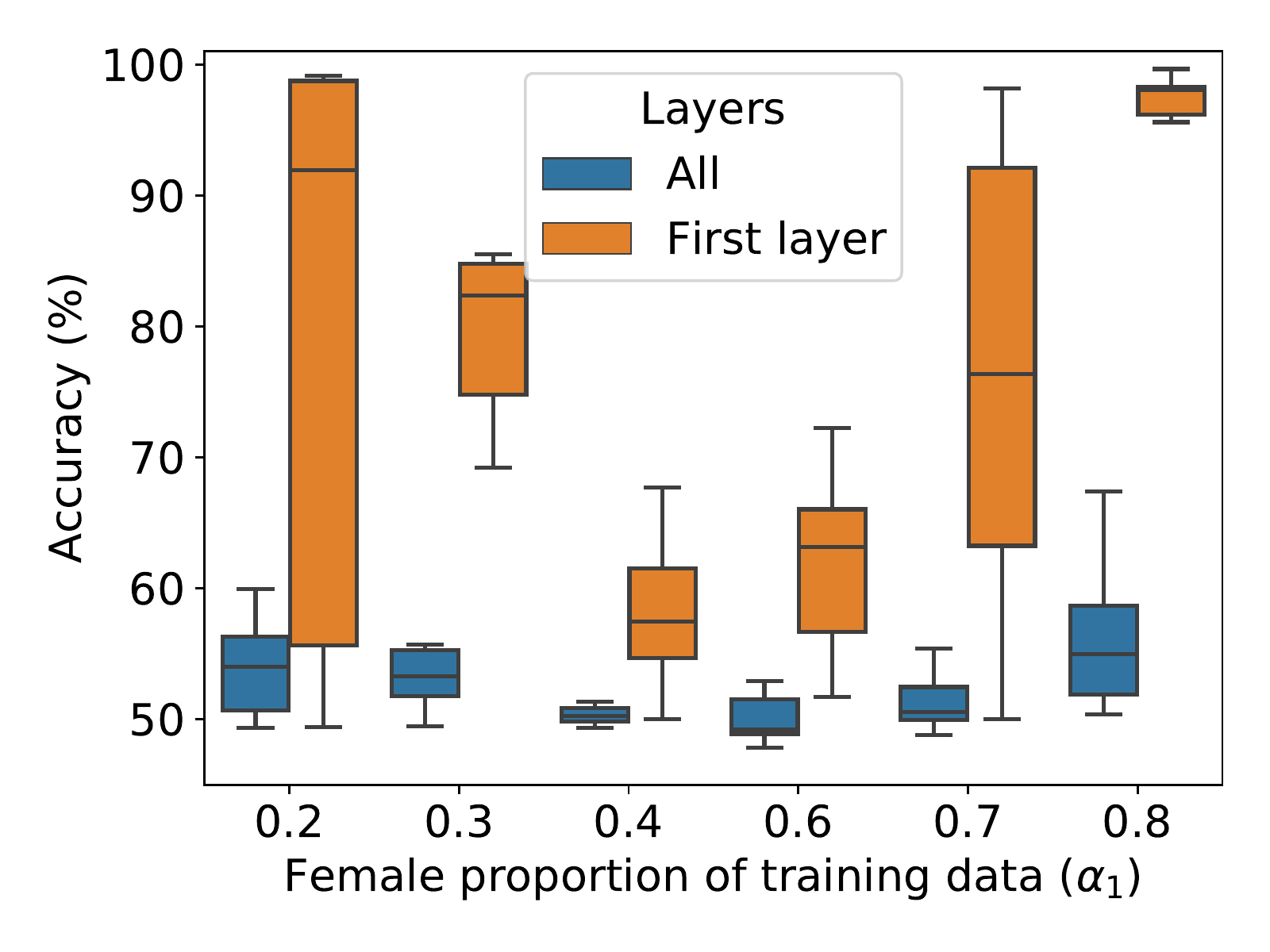}
    \caption{10 models}
    \label{fig:boneage_10}
\end{subfigure}
\hfill
\begin{subfigure}[b]{0.32\textwidth}
    \centering
    \includegraphics[width=\textwidth]{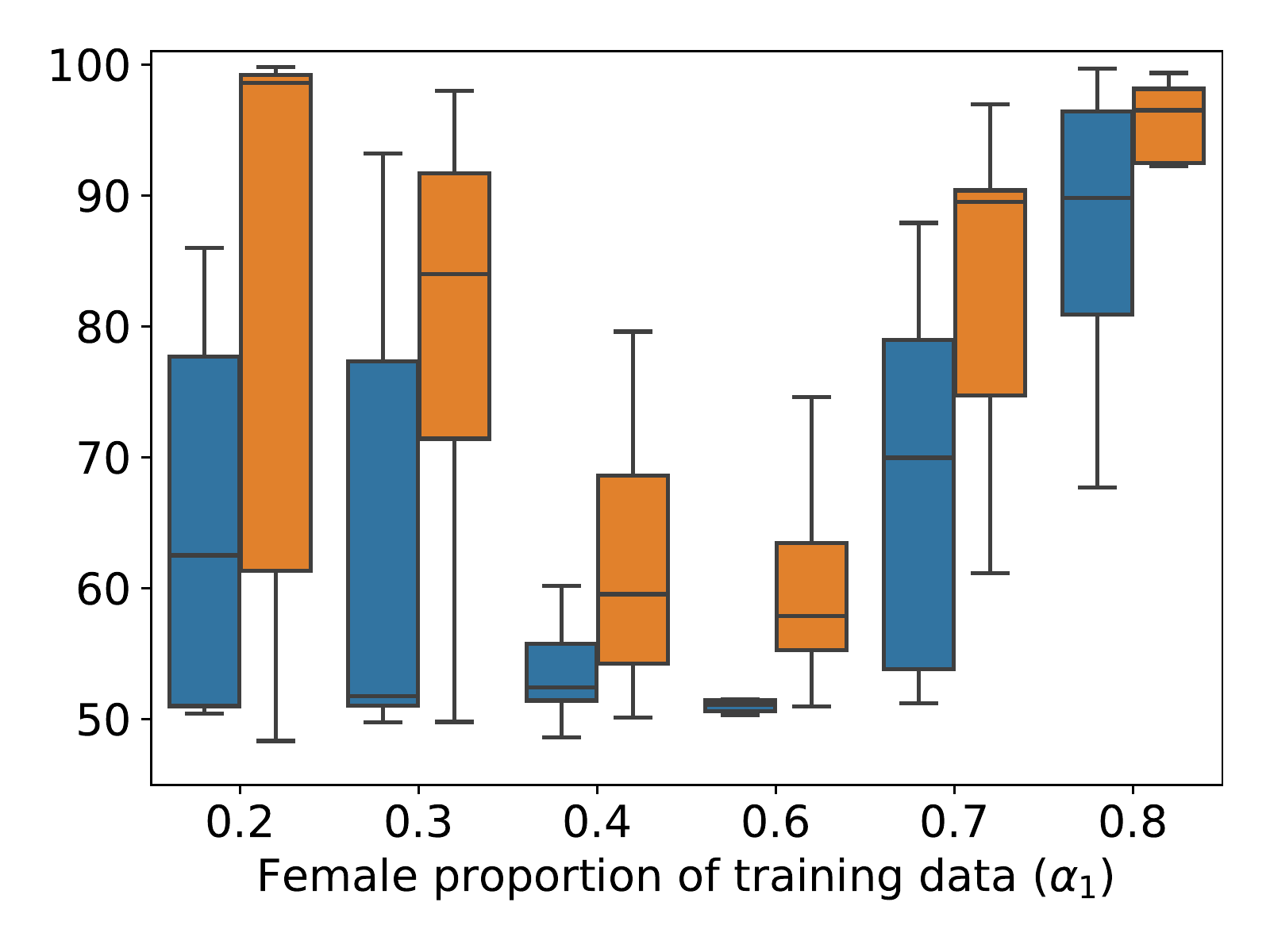}
    \caption{40 models}
    \label{fig:boneage_40}
\end{subfigure}
\hfill
\begin{subfigure}[b]{0.32\textwidth}
    \centering
    \includegraphics[width=\textwidth]{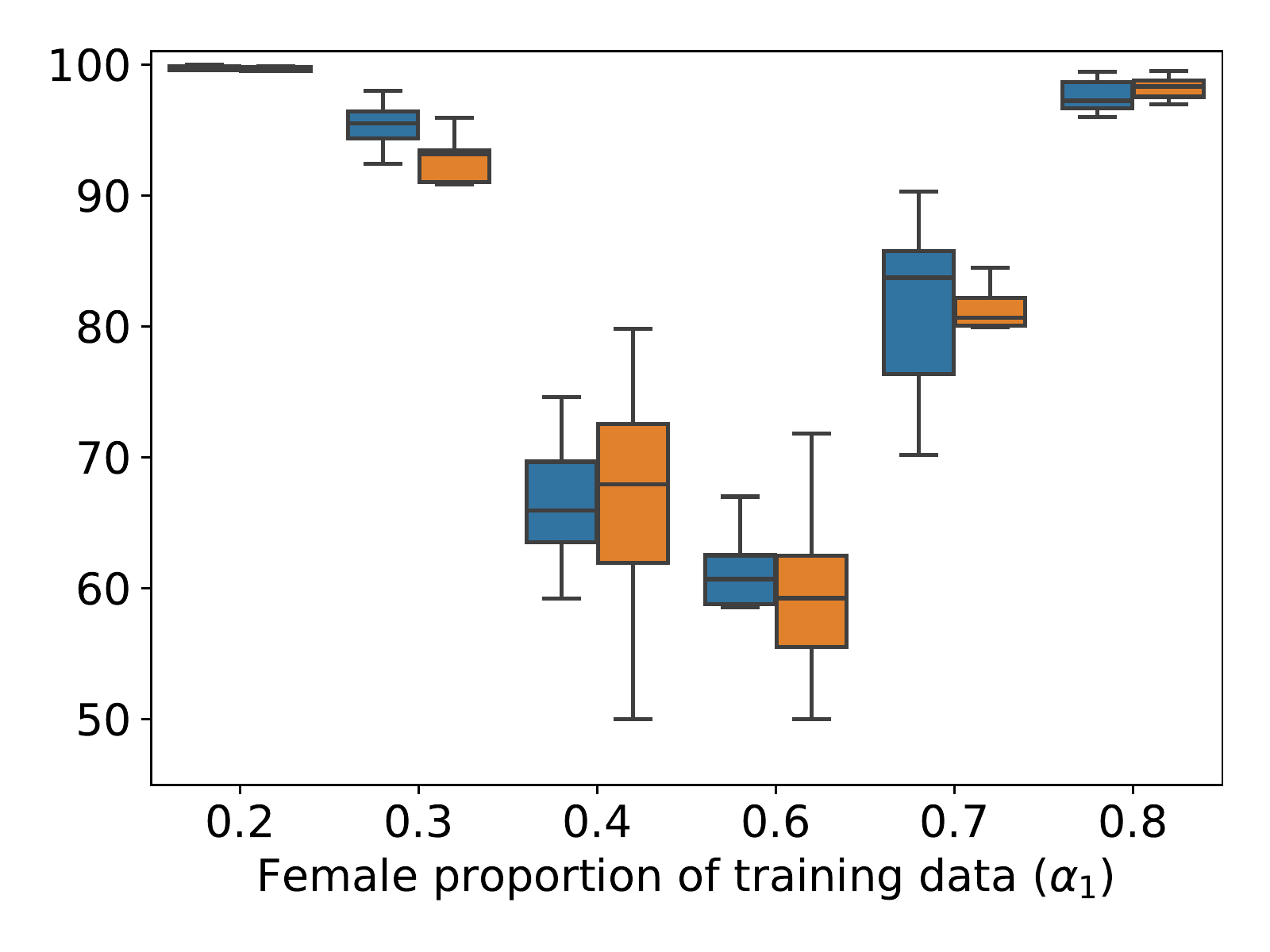}
    \caption{1600 models}
    \label{fig:boneage_1600}
\end{subfigure}
\caption{Classification accuracy for distinguishing between models with different training distributions on the \boneage\ dataset, for meta-classifiers trained with (\protect\subref{fig:boneage_10}) 10, (\protect\subref{fig:boneage_40}) 40, and (\protect\subref{fig:boneage_1600}) 1600 models. Orange box plots correspond to using parameters only from the first layer, while blue box plots correspond to using all (three) layers' parameters. Although the experiment that uses just the first-layer's parameters has much more variance, its average performance (and even the first quartile) is better than that of the version that uses all the layer's parameters.}
\label{fig:boneage_varying_n}
\end{figure*}

\shortsection{Identifying Useful Layers}\label{sec:identifying}
Let $j$ be some layer of the model for which the adversary wishes to gauge inference potential. We optimize query point $\hat{x}$ to maximize the difference in the total number of activations for layer $j$ between models trained on datasets from the two distributions:
\begin{align}
    M_j(x) & = \sum_i \mathbb{I}[(M[:j](x))_i > 0] \nonumber\\
    \hat{x} & = \arg\max_{x} \left|\sum_{i;y_i=0}M^i_j(x) - \sum_{i;y_i=1}M^i_j(x) \right|, \label{eq:identify_useful}
\end{align}
where $M[:j](x)$ refers to the activations after layer $j$ of model $M$ on input $x$. The adversary can use a set of test points to select one that maximizes the above constraint. Then, similar to the process for \ttest\  (Equation~\ref{eq:threshold}), the adversary finds a threshold on the number of activations to maximize distinguishing accuracy. By iterating through all layers and computing the corresponding accuracies, the adversary can create a ranking of layers to estimate how much information these layers can potentially leak. This process is computationally much cheaper than running a meta-classifier experiment for all layers, and can be done with as few as 20 models.
Once it has ranked all the layers, the adversary can pick the most informative ones (even just a single layer suffices in some cases) to train the meta-classifier. Since the resulting meta-classifier has fewer parameters (as it computes over fewer model layers), it can be trained using far fewer shadow models than when all network parameters are used, without having a significant impact on distinguishing accuracy.

\shortsection{Results}
To understand how well the layer-identification process correlates with meta-classifier performance, we also perform experiments where each layer's parameters are used one at a time to train the meta-classifier. We run the layer-identification process, as described in Equation~\ref{eq:identify_useful}, for all layers across datasets.
For the numbers reported in Table~\ref{tab:identify_layers}, the adversary samples data from its local test set to maximize Equation~\ref{eq:identify_useful}. Distinguishing accuracies reported in this table are on the adversary's models since it uses this ranking of layers to train a meta-classifier for its attack on the targeted model. For most cases, the layers closest to the inputs are identified as most useful. These accuracies for \celeba\ align with observations from previous experiments (Section~\ref{sec:results}) as well---for distinguishing sex ratios, the convolutional layers (until layer 5) seem to be more useful; for age, the fully-connected layers appear to be most useful. Layers of machine-learning models closer to the input are commonly associated with learning generic patterns, and later layers more abstract ones along with invariance to the given task~\cite{papyan2020prevalence}. Thus, the position of layers identified to be most useful is telling of how close the target property is to the input space or task.

\begin{figure*}[!h]
\centering
\begin{subfigure}[b]{0.32\textwidth}
\includegraphics[width=0.95\textwidth]{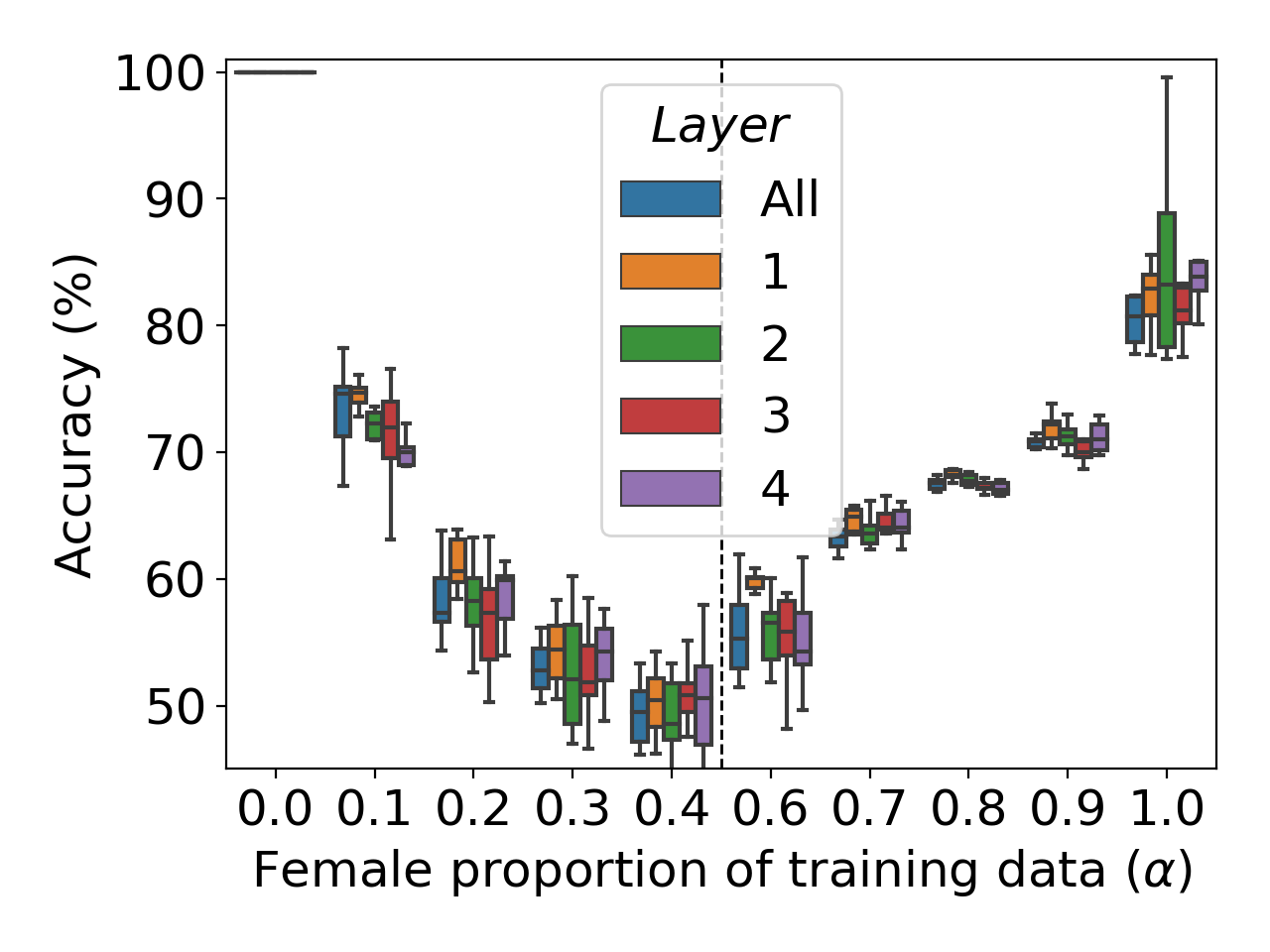}
\caption{\census, varying proportion female}
\end{subfigure}
\begin{subfigure}[b]{0.32\textwidth}
\includegraphics[width=0.95\textwidth]{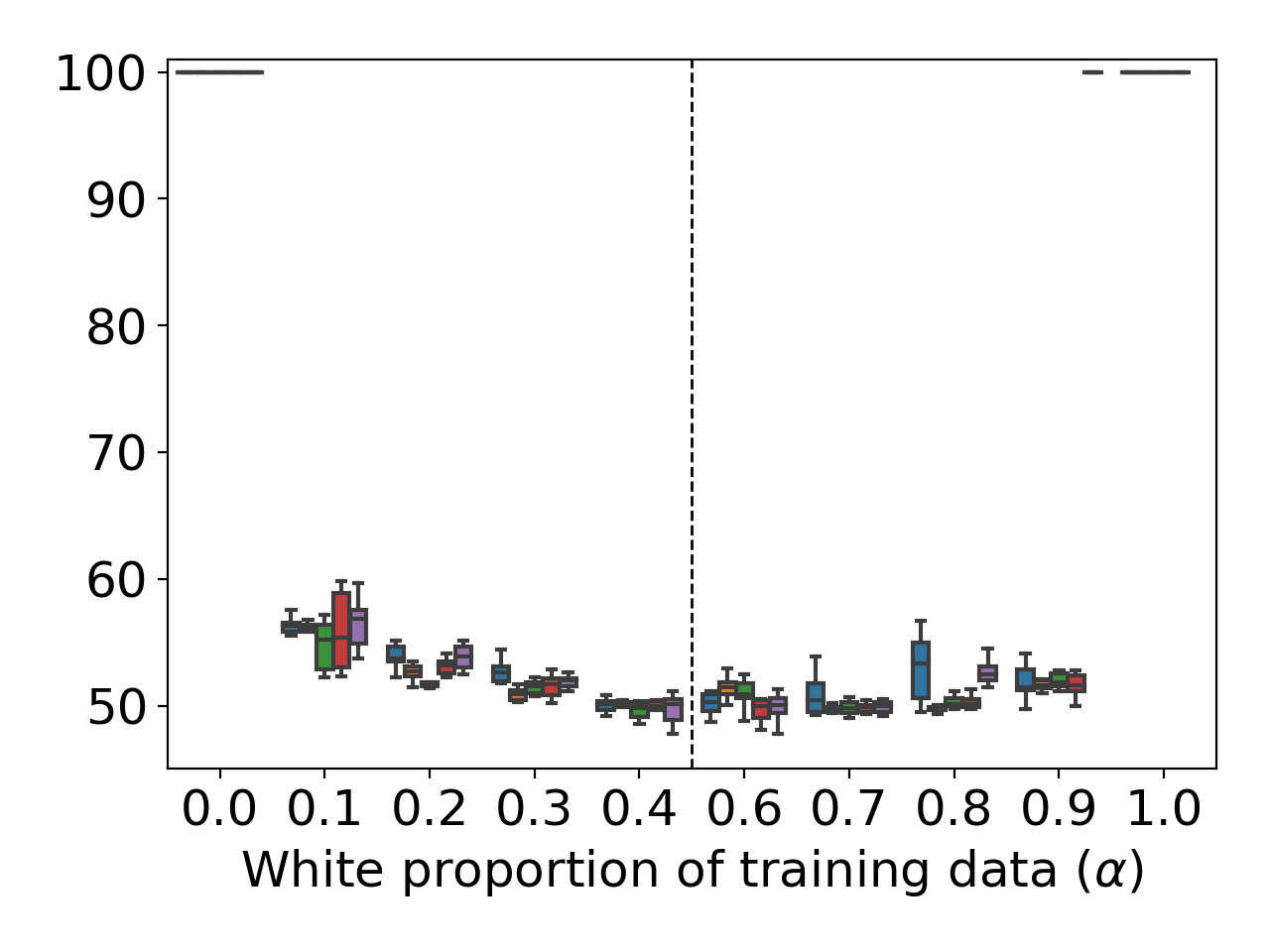}
\caption{\census, varying proportion white}
\end{subfigure}
\begin{subfigure}[b]{0.32\textwidth}
\includegraphics[width=0.95\textwidth]{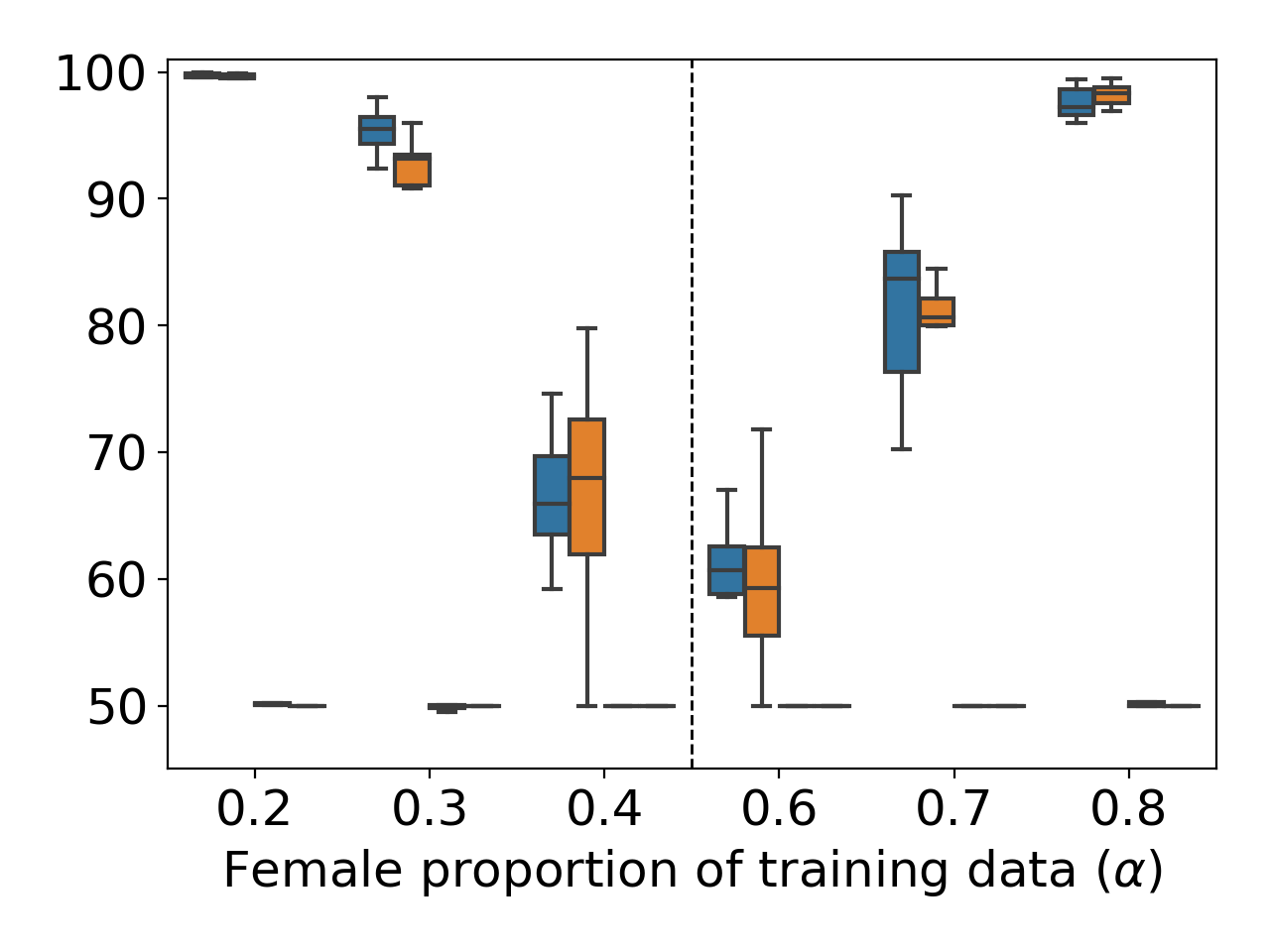}
\caption{\boneage, varying proportion female}
\end{subfigure}
\caption{Classification accuracy for distinguishing between training distributions for unseen models for on \census\ (sex: left, race: middle) and \boneage, while varying the models' layers used while training meta-classifiers. There is no clear winner in the case of \census, while the first layer seems to the most useful for the case of \boneage.}
\label{fig:census_boneage}
\end{figure*}

\begin{figure*}[!h]
\centering
\begin{subfigure}[b]{0.32\textwidth}
\includegraphics[width=0.95\textwidth]{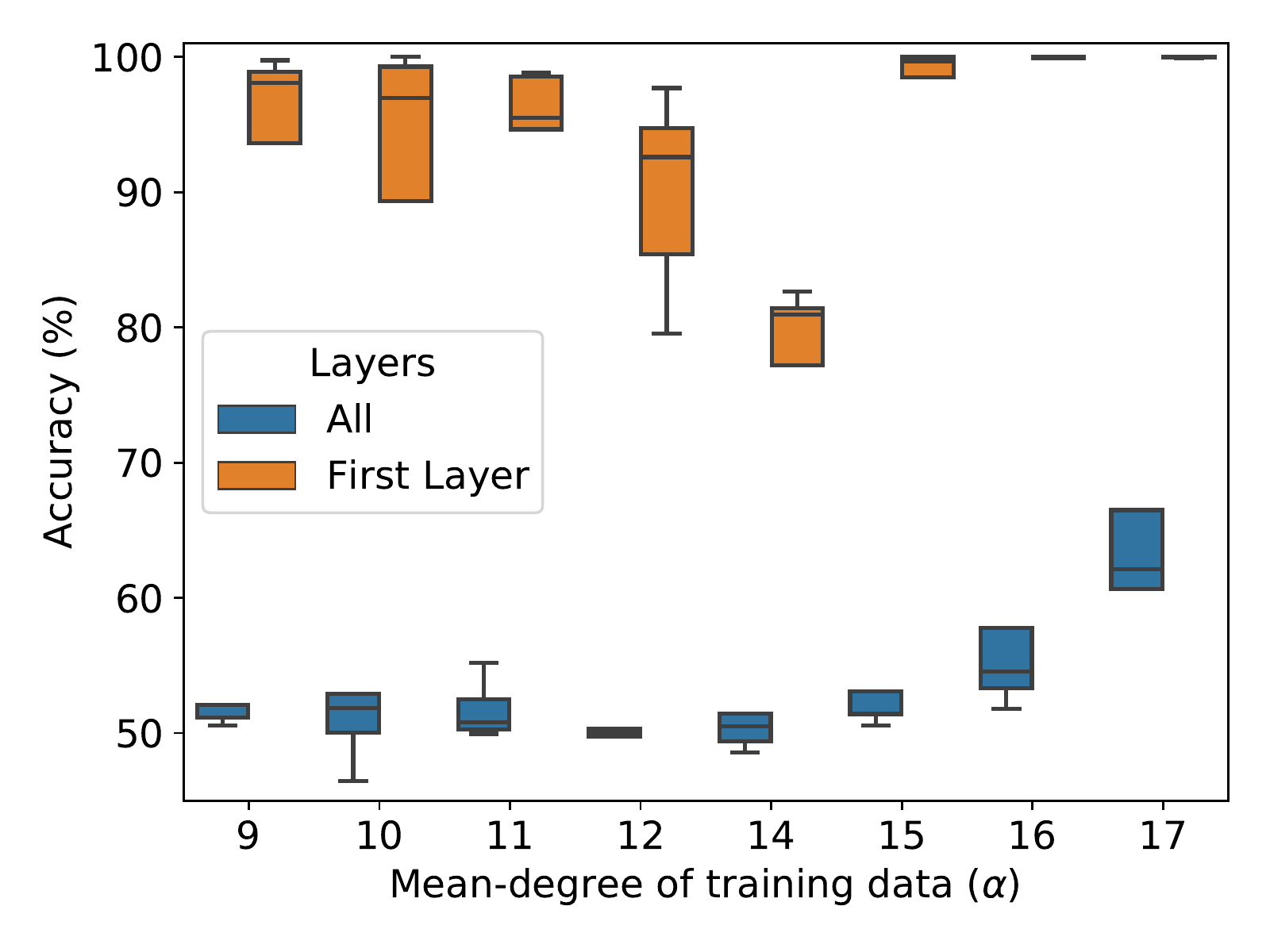}
\caption{20 models}
\end{subfigure}
\hfill
\begin{subfigure}[b]{0.32\textwidth}
\includegraphics[width=0.95\textwidth]{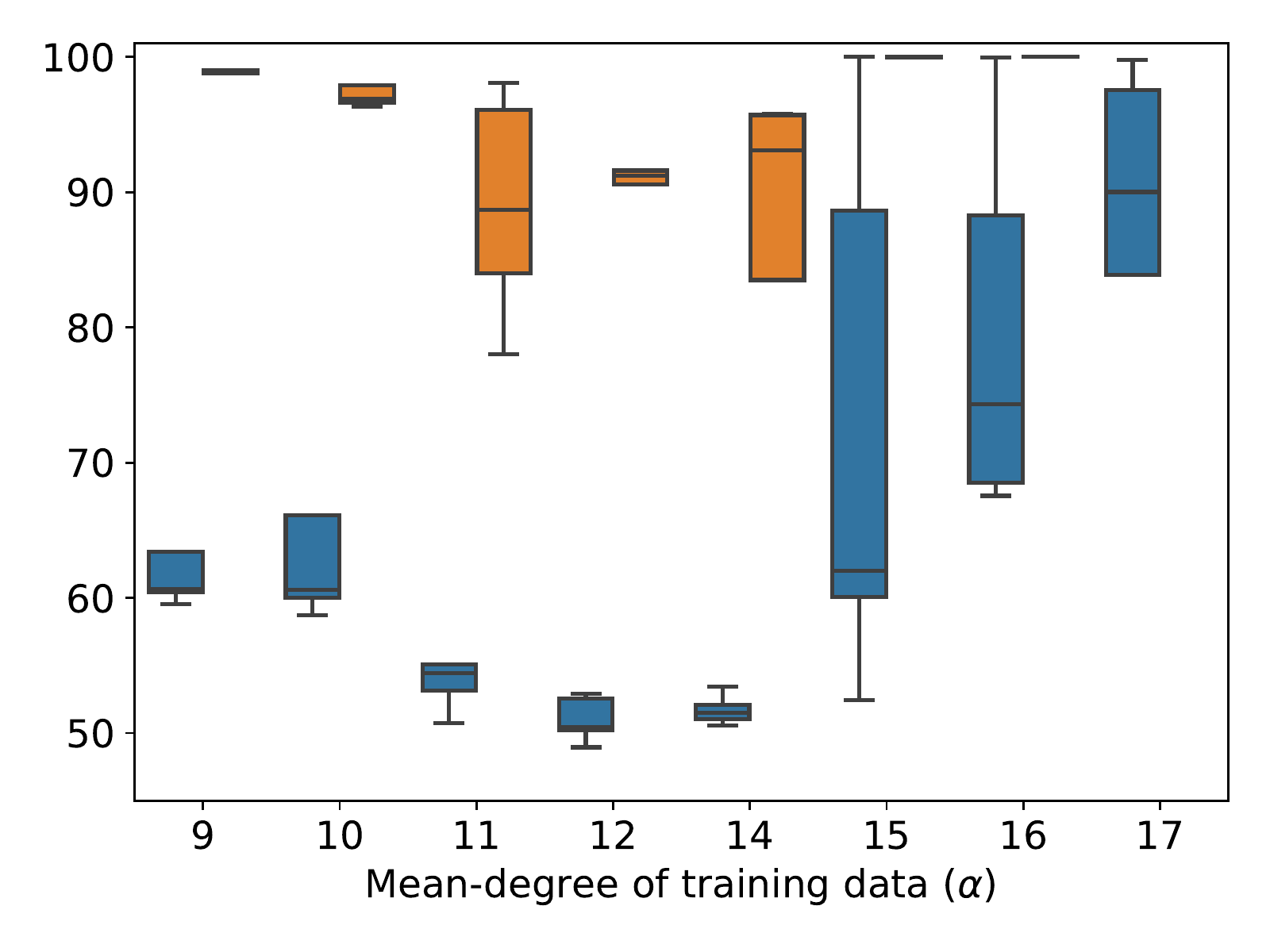}
\caption{100 models}
\end{subfigure}
\hfill
\begin{subfigure}[b]{0.32\textwidth}
\includegraphics[width=0.95\textwidth]{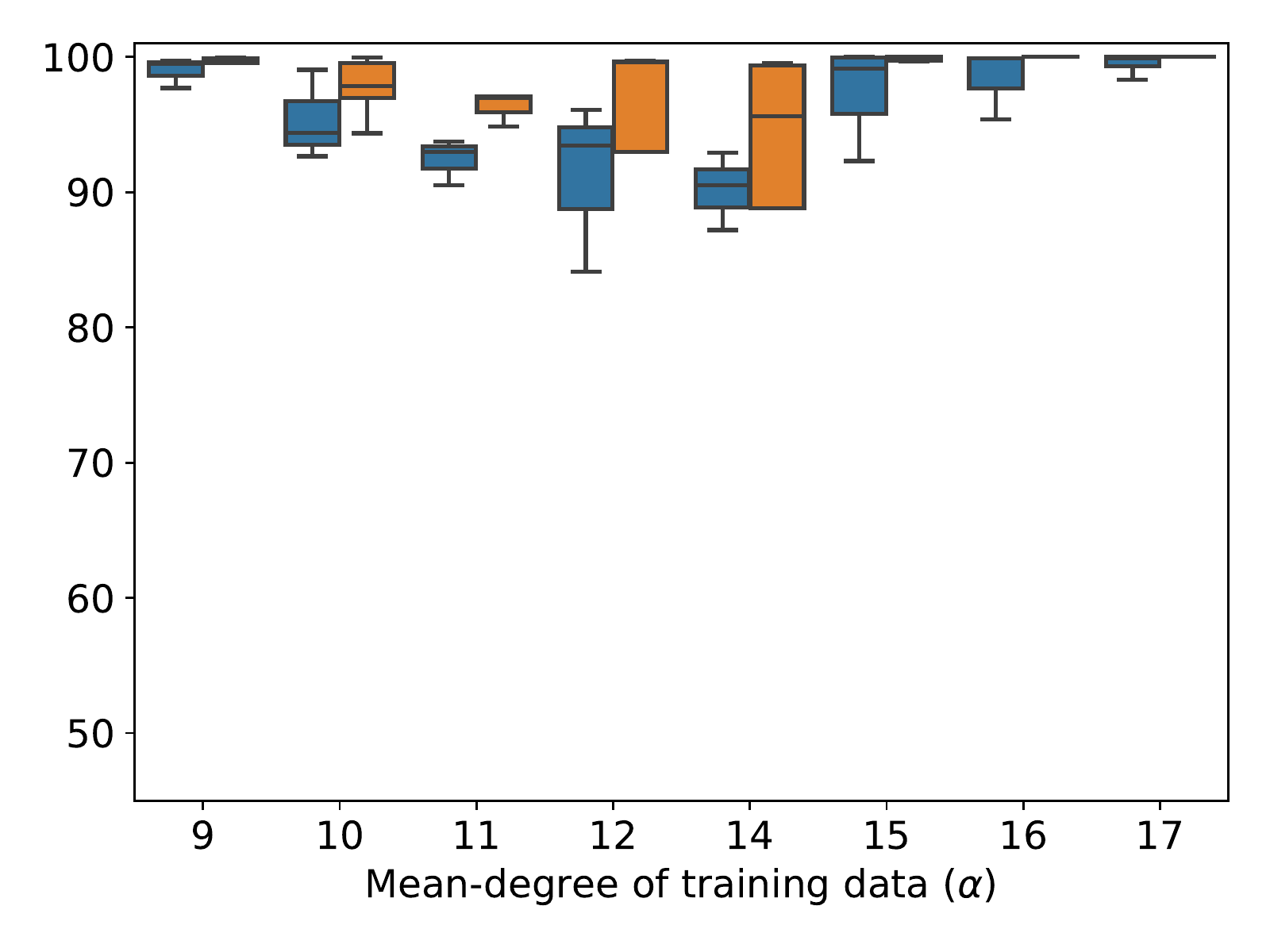}
\caption{1600 models}
\end{subfigure}
\caption{Classification accuracy for distinguishing between models with different training distributions for on the \graph\ dataset. Left to right: meta-classifiers trained using 20, 100, and 1600 (original experiment) models, respectively. Orange box plots correspond to using parameters only from the first layer, while blue box plots correspond to using all (three) layers' parameters. Using as few as 20 models is sufficient for satisfactory meta-classifier performance when the right layers are identified and used.}
\label{fig:arxiv_varying_n}
\end{figure*}

Excluding the last layer does not lead to a significant performance drop. Intuitively, layers closer to the output will capture invariance for the given task and are thus less likely to contain any helpful information that prior layers would not already capture. If the last layer reveals enough information for the attack to succeed, then a black-box attack should also be possible. Results from layer-wise meta-classifier experiments confirm how the last layer's parameters rarely appear useful for distribution inference. Using these observations, we train meta-classifiers while using parameters only from some of the layers selected on the ranking we obtain via layer-identification experiments. We observe a clear advantage of doing so across all datasets, with minimal decreases in accuracy. For instance, using just the first layer produces a meta-classifier with only 20 training models on \boneage\ (orange boxes in leftmost graph in Figure~\ref{fig:boneage_varying_n}) that performs much better than using parameters from all of the layers. In order for the meta-classifier trained on all parameters to approach the accuracy of the one-layer meta-classifier, hundreds of shadow models are needed.

For datasets like \census, all the layers seem to equally useful while for \boneage, only the first layers' parameters are useful (see Figure~\ref{fig:census_boneage}). When using the first layer's parameters, the adversary can achieve an average of $75\%$ accuracy with as few as 20 models, compared to 54\% when using the entire model.
In fact, the first layer is identified as most useful and using any other layer leads to near-random performance. 
Additionally, for larger models like those for \celeba, the adversary can pick more than one layer---using as few as three layers of the model can help lower computational resources. As observed in ablation experiments with convolutional and linear layers for \celeba\ (old people), using just the last three layers (of which two the layer-identification process identifies), the adversary can train its meta-classifiers while using significantly fewer models.
For instance, when using 100 models to train the meta-classifier, using just the fully-connected layers gives a $4\%$ absolute improvement in accuracy, along with $0.5\%$ reduction in standard deviation across experiments.


\shortersection{Graph Datasets} The layer-identification process does not work on the graph datasets. It incorrectly predicts the third layer as most useful for \graph, whereas actual performance with that layer's parameters leads to a significant performance drop. We suspect this behavior can be explained by the inherent properties of the graph data.  Intermediate activations for nodes can have complicated interactions with neighboring nodes, leading to the detection method's instability when analyzing activation values. 
These challenges on graph datasets is something that we plan to investigate in future work. Nonetheless, the fact the first two layers are useful for both graph datasets suggests a useful direction for graph-based property inference attacks. 

\end{document}